%% file: final_chapters/survey.tex
\documentclass[lettersize,journal]{IEEEtran}
\usepackage{amsmath,amsfonts}
\usepackage{algorithmic}
\usepackage{algorithm}
\usepackage{array}
\usepackage[caption=false,font=normalsize,labelfont=sf,textfont=sf]{subfig}
\usepackage{textcomp}
\usepackage{stfloats}
\usepackage{url}
\usepackage{verbatim}
\usepackage{graphicx}
\usepackage{cite}
\usepackage{booktabs,longtable,array,caption,multirow}
\usepackage[table]{xcolor}
\usepackage[colorlinks]{hyperref}
\usepackage{empheq}
\usepackage{threeparttable}

\usepackage[capitalize]{cleveref}
\crefname{section}{Sec.}{Secs.}
\Crefname{section}{Section}{Sections}

\crefname{subsection}{Sec.}{Secs.}
\Crefname{subsection}{Section}{Sections}

\crefname{subsubsection}{Sec.}{Secs.}
\Crefname{subsubsection}{Section}{Sections}

\crefname{figure}{Fig.}{Figs.}
\Crefname{figure}{Figure}{Figures}

\crefname{table}{Tab.}{Tabs.}
\Crefname{table}{Table}{Tables}

\crefname{equation}{Eq.}{Eqs.}
\Crefname{equation}{Equation}{Equations}

\usepackage[percent]{overpic}
\newcommand{\wm}{World Model}
\newcommand{\wms}{World Models}
\newcommand{\wam}{World-Action Model}
\newcommand{\wams}{World-Action Models}

\newcommand{\vlas}{Vision-Language-Action Models}

\begin{document}

\markboth{Journal of \LaTeX\ Class Files,~Vol.~14, No.~8, August~2021}%
{Shell \MakeLowercase{\textit{et al.}}: A Sample Article Using IEEEtran.cls for IEEE Journals}


\title{World-Action Models for Robot Learning and Control: A Survey}


\author{
Zuxing~Lu$^{1,*}$, 
Hongjia~Zhai$^{1,*}$, 
Guanzhi~Wang$^{2}$, 
Huajian~Zeng$^{1}$, 
Jiaqi~Yang$^{1}$, 
Jingyu~Liu$^{1}$, 
Lei~Cheng$^{1}$, 
Yuantai~Zhang$^{1}$, 
Yuheng~Qiu$^{3}$,
Zezhou~Cheng$^{4}$, 
Ivan~Laptev$^{1}$, 
Danfei~Xu$^{5}$, 
Benjamin~Riviere$^{6}$,
Giuseppe~Loianno$^{7}$, %
Eric~Xing$^{1}$, %
Xingxing~Zuo$^{1,\dagger}$\\[0.2em] 
\small
$^{1}$MBZUAI
\quad
$^{2}$Caltech
\quad
$^{3}$Amazon FAR
\quad
$^{4}$University of Virginia
\quad
$^{5}$Georgia Tech
\quad
$^{6}$New York University
\quad
$^{7}$UC Berkeley\\
\small
$^{*}$Equal contribution
\quad
$^{\dagger}$Corresponding author
}

\input{final_chapters/figs/fig1_cover_fig.tex}

\begin{abstract}
Robots operating in open environments act under partial observability, physical constraints, and dynamic task contexts. Beyond mapping observations and language instructions to actions, they must anticipate how candidate actions may affect future states and task-relevant outcomes. Recent advances in world models, video generation, and Vision-Language-Action (VLA) policies have motivated the development of World-Action Models (WAMs), which couple future world prediction with executable action generation. This survey provides a robotics-oriented review of WAMs. We clarify their scope relative to conventional world models, model-based reinforcement learning, action-conditioned video generation, and reactive VLA policies, and organize existing methods through a unified taxonomy covering representations, transition modeling, action interfaces, architectures, training pipelines, data modalities, and scaling strategies. We further review applications of WAMs in manipulation, navigation, and autonomous driving, and we summarize the datasets, benchmarks, metrics, and protocols used to evaluate WAM systems. Finally, we discuss key challenges in action alignment, world-action factorization, spatial and multi-view consistency, long-horizon memory, neural simulation for closed-loop policy learning, and efficient inference. Taken together, this survey aims to provide a concise technical foundation for integrating predictive world modeling with action generation, toward more reliable embodied robot intelligence.
Project page: \url{https://rcl-robotics.github.io/Awesome-World-Action-Models}.
\end{abstract}


\input{final_chapters/Sec-I.tex}
\input{final_chapters/Sec-II.tex}
\input{final_chapters/Sec-III.tex}

\input{final_chapters/Sec-IV.tex}
\input{final_chapters/Sec-V.tex}

\input{final_chapters/Sec-VI.tex}

\input{final_chapters/Sec-VII.tex}

\bibliographystyle{IEEEtran}
\bibliography{ref}

\end{document}

%% file: final_chapters/figs/fig1_cover_fig.tex
\twocolumn[{
\renewcommand\twocolumn[1][]{#1}
\maketitle
\begin{center}\vspace{-0.75cm}
\renewcommand{\tabcolsep}{1pt}
\begin{overpic}[width=1.\textwidth]{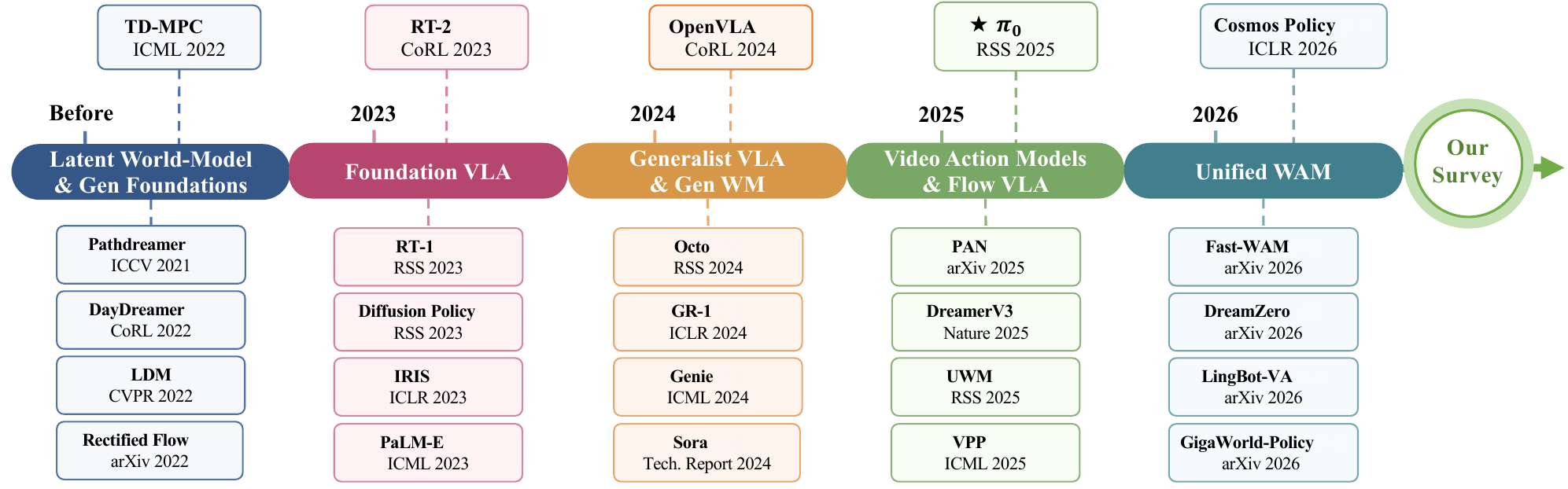}

\put (13.27,29.1) {\footnotesize{\cite{td_mpc}}} 
\put (11.83,15.1) {\scriptsize{\cite{pathdreamer}}} 
\put (11.76,11.1) {\scriptsize{\cite{daydreamer}}} 
\put (10.93,6.9) {\scriptsize{\cite{ldm}}} 
\put (12.03,2.8)  {\scriptsize{\cite{rectified_flow}}} 

\put (29.14,29.1) {\footnotesize{\cite{rt2}}} 
\put (27.96,15.1) {\scriptsize{\cite{rt1}}} 
\put (30.33,11.1) {\scriptsize{\cite{diffusion_policy}}} 
\put (27.69,6.9) {\scriptsize{\cite{iris}}} 
\put (28.59,2.8)  {\scriptsize{\cite{palme}}} 

\put (48.49,29.1) {\footnotesize{\cite{openvla}}} 
\put (45.54,15.1) {\scriptsize{\cite{octo_2023}}} 
\put (45.67,11.1) {\scriptsize{\cite{gr1}}} 
\put (45.76,6.9) {\scriptsize{\cite{genie}}} 
\put (45.42,2.8)  {\scriptsize{\cite{sora}}} 

\put (65.81,29.1) {\footnotesize{\cite{pi0}}} 
\put (63.59,15.1) {\scriptsize{\cite{pan}}} 
\put (64.89,11.1) {\scriptsize{\cite{dreamerv3}}} 
\put (63.57,6.9) {\scriptsize{\cite{uwm}}} 
\put (63.17,2.8)  {\scriptsize{\cite{vpp}}} 

\put (85.20,29.1) {\footnotesize{\cite{cosmos_policy}}} 
\put (82.48,15.1) {\scriptsize{\cite{fastwam}}} 
\put (82.57,11.1) {\scriptsize{\cite{dreamzero}}} 
\put (82.73,6.9) {\scriptsize{\cite{lingbotva}}} 
\put (83.79,2.8)  {\scriptsize{\cite{gigaworld_policy}}} 

\end{overpic}
\end{center}
\label{fig:WAM_timeline}
\small \hypertarget{fig:SLAM_timeline}{Fig. 1:} {\textbf{Evolution of World-Action Models.} The timeline summarizes representative milestones from latent world-model reinforcement learning and foundation vision-language-action models to video-action models, and recent unified World-Action Models. It highlights how predictive modeling and action generation have gradually converged toward integrated frameworks for embodied robot control.
}
\vspace{0.5cm}
}]

%% file: final_chapters/Sec-I.tex
\section{Introduction}

\IEEEPARstart{R}{obots} deployed in open environments executes long-horizon, multimodal, and instruction-driven tasks under partial observability and physical constraints. Unlike static perception systems, an embodied agent must interpret high-dimensional sensory inputs, infer human intent, reason about uncertain outcomes, and produce executable actions in a closed loop~\cite{dreamerv2,rt1,openvla,pi0,dreamzero,xu2025a0,li2025uncertainty,lingbotva}. Effective robotic autonomy, therefore, depends on coupling perception, prediction, and control so that decisions are informed by their expected consequences~\cite{world_model,tan2026towards}.


Classical robotics has addressed this problem through modular pipelines that decompose autonomy into perception, state estimation, planning, and low-level control~\cite{thrun2005probabilistic,Dellaert_2017,barfoot2024state,silver2010monte,sugihara2002real}. This modularity allows individual components to be independently designed, validated, and replaced, while exposing intermediate representations that facilitate interpretation, diagnosis, and systematic debugging. Such systems remain effective when task objectives, dynamics, and environmental structure can be specified with sufficient accuracy. However, their limitations become more pronounced in open-world manipulation, mobile service robotics, and multitask embodied interaction~\cite{pi0_5,wang2023dexgraspnet,gr2,daydreamer,worldvla}, where hand-designed representations are brittle under distribution shift, module interfaces can propagate errors, and task-specific models are difficult to scale across scenes, embodiments, and instructions. These considerations have motivated learning-based policies trained on large-scale and heterogeneous data.
These considerations have motivated learning-based policies trained on large-scale and heterogeneous data.

Foundation models have accelerated this shift. \vlas{} (VLAs) map visual observations and language instructions to robot actions or action chunks within a unified policy. Representative systems, including RT-1~\cite{rt1}, RT-2~\cite{rt2}, RoboFlamingo~\cite{li2023vision}, Octo~\cite{octo_2023}, and OpenVLA~\cite{openvla}, show that large-scale pretraining and cross-task data aggregation can improve instruction following, policy generalization, and transfer across tasks and embodiments. By aligning visual observations, linguistic goals, and motor commands, VLAs provide a scalable interface between human intent and robot execution~\cite{Fei2025LIBEROPlusIndepth,spatialforcing2025}. They have also shifted robot learning from narrow task-specific policies toward generalist policies trained on heterogeneous robot data and, in some cases, web-scale vision-language representations~\cite{Guzey2025DexteritySmart,Punamiya2025EgoBridgeDomain,Wang2026HumanXAgile,Zhang2026CLAPContrastive}.

Despite this progress, most VLA policies remain primarily reactive. They are typically trained to map the current observation, a short history, and a task instruction to the next action, without explicitly modeling how alternative actions would change future observations, physical states, or task progress~\cite{openvla,pi0,pi0_5,rt2}. This limitation is critical in long-horizon manipulation, delayed-effect tasks, occluded scenes, and contact-rich interactions~\cite{Fei2025LIBEROPlusIndepth,robohorizon,mind_v}, where an action that appears locally plausible may prevent later subgoals from becoming reachable. Therefore, world models address a complementary problem by learning how observations, states, rewards, or latent variables evolve as a function of context and actions~\cite{xing2025critiques}. Rooted in model-based reinforcement learning and predictive coding, this idea has been instantiated in systems such as Dreamer~\cite{dreamer} and TD-MPC~\cite{td_mpc}, as well as in recent latent video prediction, action-conditioned generation, and embodied simulation models~\cite{iris,diamond,polygrad,adaworld}. For robotics, world models are attractive because they support imagination, planning, counterfactual evaluation, data-efficient learning, and uncertainty-aware control~\cite{robotic_wm,iws,worldgym,ctrl_world}.

A predictive world model alone, however, is not sufficient for effective robot control. In many model-based formulations, the learned dynamics model, planner, and policy are trained or deployed as separate components~\cite{td_mpc2,dreamerv3,pathdreamer,occworld}. This separation can create a mismatch between predictive fidelity and control utility: visually plausible futures are not necessarily actionable, and accurate short-horizon prediction does not guarantee robust long-horizon behavior~\cite{seer,xing2025critiques}. Explicit planning over learned dynamics can also be computationally expensive, sensitive to model bias, and difficult to combine with high-frequency control and multimodal task conditioning.

These limitations motivate World-Action Models (WAMs)~\cite{uwm,vla_wm,gigabrain0,pi0_7,fastwam,dreamzero}. In this survey, we use WAMs to denote models that couple predictive world representations with executable action generation, either by jointly predicting future states and actions or by predicting future states and then inferring actions through an inverse-dynamics interface. Under this view, future-state imagination, latent dynamics, action decoding, and policy learning are connected within a shared learning or inference process~\cite{world_vla_loop,eva_vwm,world4rl}. Compared with standard VLA policies, WAMs introduce predictive structure for reasoning about action consequences~\cite{vidman,vpp,cosmos_policy}. Compared with conventional model-based reinforcement learning, WAMs increasingly incorporate multimodal foundation-model conditioning, large-scale offline robot data, and amortized action generation. This perspective provides a common lens for latent-imagination policies~\cite{daydreamer,dreamerv3}, action-conditioned video generators~\cite{gaia_1,genie_env,inspatio_world}, diffusion-based trajectory and policy models~\cite{polygrad}, unified video-language-action models~\cite{worldvla,pi0,ma2026dit4dit}, and robotic systems that use predicted futures as part of closed-loop control~\cite{vidman,vpp,cosmos_policy}.

These observations motivate a robotics-oriented survey that goes beyond cataloging individual methods. First, the relevant literature is distributed across video generation, embodied foundation models, VLA policies, model-based reinforcement learning, and robot imitation learning, where closely related ideas are often introduced under different names. Second, existing methods differ along several coupled design dimensions, including representation space, prediction target, action interface, transition formulation, architecture, training objective, and control usage. Third, current evaluation protocols often measure predictive quality, imitation accuracy, planning performance, and real-robot robustness separately, leaving open whether imagined futures are physically feasible, temporally consistent, and ultimately effective in improving closed-loop control. A unified treatment is therefore needed to clarify the scope, design choices, and evaluation criteria of WAMs.

To address these gaps, this article provides a robotics-oriented survey of WAMs for predictive robot control. We clarify their conceptual scope, organize existing methods through a multi-axis taxonomy, compare their roles across application domains, and summarize evaluation protocols, limitations, and future research directions. Our goal is to establish a structured view of how world prediction and action generation can be coupled to support scalable, reliable, and deployable robot intelligence.
The main contributions of this survey are summarized as follows:
\begin{itemize}
    \item \textbf{Definition and scope.}
    We define WAMs as predictive control models that connect future-state modeling with executable action generation, and clarify their relationship to conventional world models, model-based reinforcement learning, video generation, and reactive VLA policies.
    \item \textbf{Taxonomy of model design.}
    We organize existing methods by model components, transition modeling paradigms, action interfaces, architectural structures, training pipelines, data modalities, and scaling strategies, covering latent dynamics models, video-action models, diffusion and flow-based policies, inverse-dynamics systems, and unified multimodal architectures.
    \item \textbf{Cross-domain robotics analysis.}
    We review WAMs in robot manipulation, navigation, autonomous driving, and generalist embodied robotics, emphasizing whether prediction is used for representation learning, look-ahead planning, synthetic data generation, or imagined policy optimization.
    \item \textbf{Evaluation and open challenges.}
    We summarize datasets, benchmarks, metrics, and evaluation protocols, and discuss open challenges in action grounding, spatial consistency, neural simulation, real-time inference, embodiment transfer, and safe deployment.
\end{itemize}

The remainder of this article is organized as follows. \cref{sec:background} recaps the main technical foundations of WAMs and presents a unified formulation that jointly models future observations and actions. \cref{sec:architecture} introduces the taxonomy of WAM architectures and learning paradigms, covering model components, transition modeling, architectural structures, training pipelines, and data scaling. \cref{sec:applications} surveys applications in manipulation, navigation, autonomous driving, and generalist embodied robotics. \cref{sec:datasets} summarizes datasets, benchmarks, and evaluation metrics, with emphasis on whether predicted futures are useful for control. \cref{sec:challenges} discusses open challenges and future research directions, and \cref{sec:conclusion} concludes the survey.

%% file: final_chapters/Sec-II.tex
\section{Background: From \wms{} to \wams{}}
\label{sec:background}


WAMs build on the design principles of WMs and VLAs, but treat future-state prediction and executable action generation as interrelated components of a unified framework.
This section clarifies their distinctions in \cref{subsec:brief-definition}, reviews foundational threads in \cref{subsec:foundational-threads}, and connects them to a unified generative view in \cref{subsec:threads-to-wam}.


\subsection{Definition}
\label{subsec:brief-definition}

We consider an embodied agent interacting with its environment and formalize the interaction as a partially observable Markov decision process (POMDP)~\cite{kaelbling1998planning}. The POMDP components are denoted as $\langle\mathcal{S}, \mathcal{A}, \mathcal{T}, \mathcal{R}, \Omega, \mathcal{O} \rangle$, 
where $\mathcal{S}$, $\mathcal{A}$, $\mathcal{T}$, and $\mathcal{R}$ describe a Markov decision process, representing the state space, action space, transition function, and reward function, respectively; 
Moreover, $\Omega$ denotes the set of observations available to the agent, which may include visual observations, robotic-arm poses, and action histories, among other sensory and proprioceptive inputs.
\( \mathcal{O}: \mathcal{S} \times \mathcal{A} \to \Pi(\Omega) \) is the observation function, which gives, for each action and resulting state, a probability distribution over possible observations.
We further define the latent state space $\mathcal{Z}$ as a compressed, low-dimensional encoding space of interaction histories sufficient for predicting future dynamics; the latent action space $\mathcal{U}$ as a compact representation space of behaviors bridging high-level commands and low-level motor control, and the language instruction $\ell$ as a natural language description of the task that conditions the agent's behaviors.



\textbf{\wms{} (WMs)}~\cite{world_model} are predictive models of environment dynamics that enable agents to perform reasoning and prediction without direct interaction. 
Typically, a world model is defined by a latent state space $\mathcal{Z}$, an action space $\mathcal{A}$, and a transition function $f_\textit{WM}$ that maps past states and actions to the next state:
\begin{equation}
    \mathbf{\hat{z}}_{t+1} = f_\textit{WM}(\mathbf{z}_{t}, \mathbf{a}_{t}).
\end{equation}

\textbf{\vlas{} (VLAs)}~\cite{openvla,rt2} serve as control and decision-making frameworks that focus on generating action sequences directly from visual inputs and language instructions. 
Conditioned on the current historical observation $\mathbf{o}_{<t}$ \footnote{For notational consistency between observation and prediction sequences,  historical observation indices are shifted backward by one step.} and language instruction $\ell$, the VLA is a model that directly predicts the future action chunk $\mathbf{a}_{t:t+H}$ over a horizon of $H$ steps without explicitly modeling future state trajectories:
\begin{equation}
    \mathbf{\hat{a}}_{t:t+H} = f_\textit{VLA}(\mathbf{o}_{<t}, \ell).
\end{equation}


\textbf{\wams{} (WAMs)}~\cite{dreamzero,uwm} serve as predictive control and decision-making frameworks that focus on jointly generating future state trajectories and action sequences from visual inputs and language instructions.
Conditioned on the observation history $\mathbf{o}_{<t}$ and language instruction $\ell$, a WAM jointly predicts the future state trajectory $\mathbf{o}_{t:t+H}$ and action chunk $\mathbf{a}_{t:t+H}$ over a horizon of $H$ steps, explicitly modeling how actions affect future states:
\begin{equation}
    \mathbf{\hat{o}}_{t:t+H}, \mathbf{\hat{a}}_{t:t+H}
    = f_{\textit{WAM}}(\mathbf{o}_{<t}, \ell).
\end{equation}

\subsection{Foundational Research Threads}
\label{subsec:foundational-threads}

\begin{figure*}
    \centering
    \includegraphics[width=0.98\linewidth]{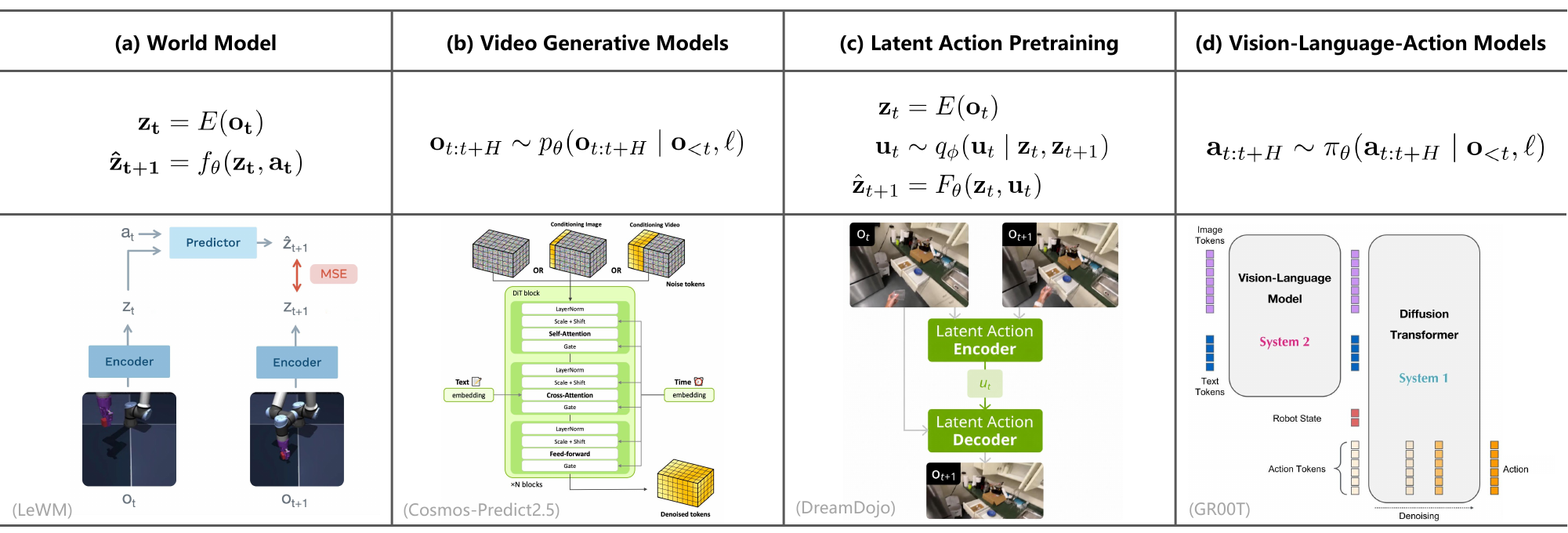}
    \caption{Conceptual overview of four foundational threads behind WAMs: (a) world models learn latent dynamics by predicting the next state from current state and action; (b) video generative models predict future visual observations conditioned on visual history and language instruction; (c) latent action pretraining infers latent actions from consecutive frames and uses them to model latent transitions; and (d) VLAs generate future action sequences directly from state and language instructions. 
    (The third-row illustrations are adapted from the corresponding papers cited in each thread.) 
    }
    \label{fig:wam_background}
\end{figure*}

As summarized in \cref{fig:wam_background}, these research threads respectively provide dynamics modeling, visual prediction, latent action inference, and direct action generation.

\subsubsection{\wm{} and Model-based Reinforcement Learning}
\label{subsec:mbrl}

Reinforcement learning (RL) is commonly divided into model-free (MFRL) and model-based (MBRL) paradigms. Unlike MFRL, which learns policies directly from interaction, MBRL leverages a dynamics model for planning and policy optimization. A representative example is PlaNet~\cite{hafner2019learning}, which introduces the Recurrent State-Space Model (RSSM) to model latent dynamics with deterministic and stochastic states:
\begin{equation}
\begin{aligned}
    &\text{Deterministic state model:} & \mathbf{h}_t &= f(\mathbf{h}_{t-1}, \mathbf{s}_{t-1}, \mathbf{a}_{t-1}) \\
    &\text{Stochastic state model:} & \mathbf{s}_t &\sim p(\mathbf{s}_t \mid \mathbf{h}_t) \\
    &\text{Observation model:} & \mathbf{o}_t &\sim p(\mathbf{o}_t \mid \mathbf{h}_t, \mathbf{s}_t) \\
    &\text{Reward model:} & r_t &\sim p(r_t \mid \mathbf{h}_t, \mathbf{s}_t).
\end{aligned}
\end{equation}
where $\mathbf{h}_t$ and $\mathbf{s}_t$ denote the deterministic and stochastic components of the latent state, respectively; $\mathbf{a}_t$, $\mathbf{o}_t$, and $r_t$ denote the action, observation, and reward; and $f$ is an RNN-based transition function.
Subsequent world models, including explicit approaches~\cite{dreamer,dreamerv2,dreamerv3} that use a decoder and implicit approaches~\cite{td_mpc,td_mpc2,dino_wm,lewm} that use latent space regularization, learn latent dynamics through
\begin{equation}
\begin{aligned}
\mathbf{z}_t&=E(\mathbf{o}_t),\\
\mathbf{\hat{z}}_{t+1} &= f_\theta(\mathbf{z}_t,\mathbf{a}_t),
\end{aligned}
\end{equation}
where $\hat{\mathbf{z}}_{t+1}$ denotes the predicted latent state and $E$ is the encoder,
and $f_\theta$, equivalently denoted as $f_\textit{WM}$, is a neural world model for predicting latent dynamics.

\subsubsection{Video Generative Models}
\label{subsec:vgm}

Video Generative Models (VGMs)~\cite{wan,cosmos} learn to generate future visual observations from text, images, or video history, providing powerful visual priors for long-horizon prediction and planning. In embodied intelligence, they are better viewed as visual predictive planners rather than explicit dynamics models:
\begin{equation}
    \mathbf{o}_{t:t+H}
    \sim p_{\theta}\!\left(
        \mathbf{o}_{t:t+H}
        \mid
        \mathbf{o}_{< t}, \ell
    \right),
\end{equation}
where $p_\theta$ denotes the conditional distribution parameterized by the VGM with learnable parameters $\theta$, from which future observations are generated given the observation history and language instruction.

\subsubsection{Latent Action Pre-training}
\label{subsec:lapa}

Latent Action Pretraining (LAPA)~\cite{lapa} addresses the lack of action labels in internet videos by learning latent actions from visual state transitions and using them for downstream policy learning. The core idea is to encode motion dynamics into a structured latent action space that is transferable across domains and embodiments:
\begin{equation}
\begin{aligned}
    \mathbf{z}_t &= E(\mathbf{o}_t), \\
    \mathbf{u}_t &\sim q_\phi(\mathbf{u}_t \mid \mathbf{z}_t, \mathbf{z}_{t+1}), \\
    \hat{\mathbf{z}}_{t+1} &= F_\theta(\mathbf{z}_t, \mathbf{u}_t),
\end{aligned}
\end{equation}
where $\mathbf{u}_t \in \mathcal{U}$ denotes the latent action inferred from consecutive observations $\mathbf{o}_t$ and $\mathbf{o}_{t+1}$, $q_\phi$ denotes the latent action encoder parameterized by $\phi$, and $F_\theta$ denotes the forward dynamics model parameterized by $\theta$.

\subsubsection{\vlas{}}
\label{subsec:vla}

\vlas{} (VLAs) directly map visual observations and language instructions to actions, forming a unified framework for embodied decision-making. Existing VLAs mainly adopt either autoregressive sequence modeling~\cite{rt2,openvla}
\begin{equation}
    \pi_\theta(\mathbf{a}_{t:t+H} \mid \mathbf{o}_{<t}, \ell)
    = \prod_{h=0}^{H-1} \pi_\theta(\mathbf{a}_{t+h} \mid \mathbf{o}_{<t}, \ell, \mathbf{a}_{t:t+h}),
\end{equation}
or flow matching~\cite{pi0,pi0_5,gr00tn1_2025}
\begin{equation}
    \mathbf{a}^{\tau}_{t:t+H}
    = (1-\tau)\boldsymbol{\epsilon}
    + \tau \mathbf{a}^{1}_{t:t+H},
    \quad \tau \sim \mathcal{U}(0,1),
\end{equation}
to model action distributions conditioned on visual and language inputs, where \(\mathbf{a}^{1}_{t:t+H}\) is clean action and \(\tau\) controls the noise level.

\subsection{From These Threads to \wams{}}
\label{subsec:threads-to-wam}

The four threads above provide complementary foundations for WAMs. MBRL focuses on learning environment dynamics for planning and control; VGM enables future-scene prediction from visual observations; LAPA introduces latent action learning from action-free videos; and VLA provides an end-to-end framework for generating actions from visual and language inputs. WAMs unify these ideas within a conditional generative framework that jointly models future observations and the actions responsible for them, bridging visual prediction and decision-making.

More concretely, the generative process of WAMs can be decomposed into four components~\cite{uwm}: policy generation, forward dynamics, inverse dynamics, and visual planning. Given historical observations $\mathbf{o}_{< t}$ and a language instruction $\ell$, these components can be formalized as:
\begin{align}
    &\text{Policy:}
    && p(\mathbf{a}_{t:t+H} \mid \mathbf{o}_{< t}, \ell),
    \label{eq:wam_policy} \\
    &\text{Forward Dynamics:}
    && p(\mathbf{o}_{t:t+H} \mid \mathbf{o}_{< t}, \mathbf{a}_{t:t+H}),
    \label{eq:wam_fwd} \\
    &\text{Inverse Dynamics:}
    && p(\mathbf{a}_{t:t+H} \mid \mathbf{o}_{< t}, \mathbf{o}_{t:t+H}),
    \label{eq:wam_inv} \\
    &\text{Visual Planning:}
    && p(\mathbf{o}_{t:t+H} \mid \mathbf{o}_{< t}, \ell).
    \label{eq:wam_vis}
\end{align}

In practice, WAMs often optimize multiple components jointly. From the perspective of action generation, existing methods mainly follow two pathways. The first jointly models future observations and actions:
\begin{equation}
    \label{eq:wam_joint_prediction}
    p_{\text{joint}}(\mathbf{o}_{t:t+H}, \mathbf{a}_{t:t+H}
    \mid \mathbf{o}_{< t}, \ell).
\end{equation}

The second adopts a plan-then-act strategy, where a future visual plan is first generated and actions are then inferred through inverse dynamics (IDM):
\begin{equation}
    \label{eq:wam_plan_than_act}
    p_{\text{plan}}(\hat{\mathbf{o}}_{t:t+H} \mid \mathbf{o}_{< t}, \ell)\,
    \cdot p_{\text{idm}}(\mathbf{a}_{t:t+H} \mid \hat{\mathbf{o}}_{t:t+H}, \mathbf{o}_{< t}).
\end{equation}

These two formulations form the basis of the transition-modeling paradigms discussed in \cref{sec:wam_paradigms}.

%% file: final_chapters/Sec-III.tex
\begin{figure*}[t]
    \centering
    \includegraphics[width=0.98\linewidth]{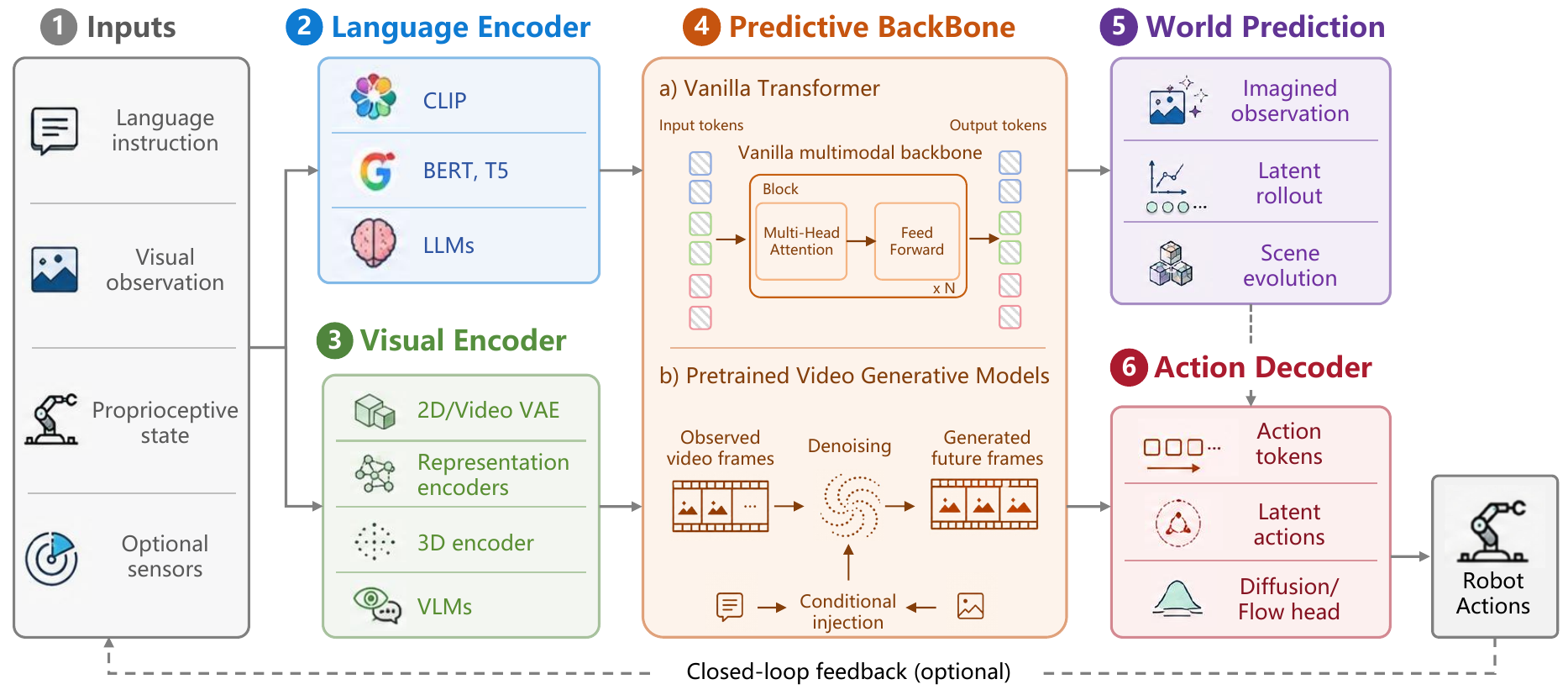}
    \caption{Component-level taxonomy for \wams{}. The architecture can be viewed through four functional modules: a language interface that provides task conditioning, a visual encoder that defines the predictive state, a predictive backbone that carries dynamics, and an action decoder that converts predicted futures into executable controls.}
    \label{fig:wam_components_taxonomy}
\end{figure*}

\section{Architecture Design \& Model Taxonomy}
\label{sec:architecture}

This section organizes WAMs along five complementary axes: core components for encoding language, observations, and actions in \cref{sec:wam_components}; transition paradigms coupling prediction with action generation in \cref{sec:wam_paradigms}; structural architectures routing multimodal tokens in \cref{sec:wam_architecture}; policy training pipelines from pre-training to post-training in \cref{sec:wam_training}; and data modality and scaling laws for effective world modeling in \cref{sec:data_modality_scaling}.

\subsection{Model Components}
\label{sec:wam_components}


WAMs transform contextual and visual inputs into actions through four functional modules: language encoder, visual encoder, action decoder, and model backbone. As summarized in \cref{fig:wam_components_taxonomy}, this decomposition separates general representation from task-specific action mechanisms, providing a consistent basis for architecture comparison.

\subsubsection{Language Encoder}
In WAMs, language encoders map instructions into conditioning representations, and existing approaches typically fall into the following three categories: 

\begin{itemize}
\item \textbf{Global Text Encoder:} CLIP~\cite{clip} learns global semantic representations through large-scale image--text contrastive learning, providing  open-vocabulary visual grounding.

\item \textbf{Dense Sequence Encoder:} T5~\cite{t5} produces dense token-level representations via span-corruption pre-training, capturing fine-grained contextual semantics for detailed language conditioning.

\item \textbf{Autoregressive LLMs:} LLMs~\cite{gpt, llama} acquire rich world knowledge and reasoning abilities through next-token prediction, enabling task understanding and long-horizon instruction following.
\end{itemize}

\subsubsection{Visual Encoder}
The visual encoder maps high-dimensional observations into compact latent representations while preserving semantic, spatial, and temporal information. Existing WAMs mainly adopt the following paradigms:

\begin{itemize}
\item \textbf{2D VAE:} Widely used in latent diffusion models~\cite{ldm,svd}, 2D VAEs compress image frames into compact latent codes, providing efficient spatial representation but limited temporal consistency.

\item \textbf{Video VAE:} Video-native VAEs~\cite{wan,yang2025cogvideox} jointly encode spatial and temporal information, producing structured latent dynamics that facilitate future video prediction.

\item \textbf{Representation Encoders:} Pretrained vision models such as DINO~\cite{dino,dinov2,dinov3}, CLIP~\cite{clip}, SigLIP~\cite{siglip} and VJEPA~\cite{vjepa2_1} extract semantic and geometric features, serving as robust representation spaces rather than pixel-level generators.

\item \textbf{3D \& Multi-view Encoders:} 3D-aware representations, including point clouds~\cite{qi2017pointnet}, occupancy grids~\cite{occworld}, and BEV features~\cite{huang2023bevdet}, explicitly encode geometric structure for spatial reasoning and control.

\item \textbf{VLMs:} Some architectures~\cite{openvla,pi0} directly reuse pretrained VLM backbones such as Qwen-VL~\cite{qwen-vl}, Prismatic~\cite{prismatic}, and PaliGemma~\cite{paligemma}, leveraging semantically rich visual features aligned with language understanding.
\end{itemize}

\subsubsection{Model Backbone}
The model backbone captures spatiotemporal dynamics from latent states and actions. Existing WAMs mainly adopt two paradigms:

\begin{itemize}
\item \textbf{Unified Transformers From Scratch}: Models such as GR-1~\cite{gr1} process multimodal inputs with a causal Transformer, learning dynamics directly from embodied and video data without relying on pretrained temporal priors.

\item \textbf{Pretrained Video Foundation Models}: Architectures such as DreamZero~\cite{dreamzero} leverage pretrained video generators like Wan~\cite{wan} and Cosmos~\cite{cosmos} as temporal priors, shifting the focus from learning dynamics to aligning visual predictions with control.
\end{itemize}

\subsubsection{Action Decoder}
The action decoder transforms latent predictions into executable control signals. Existing approaches mainly fall into three categories:

\begin{itemize}
\item \textbf{Action Tokenizer:} The continuous action space is discretized into token vocabularies using vector quantization or action bins~\cite{vqvae,rt2,openvla}, enabling LLM-compatible autoregressive prediction and scalable cross-embodiment training, albeit with quantization-induced precision loss.

\item \textbf{Latent Action Model (LAM):} LAMs~\cite{lapa,adaworld,vipra} learn abstract action representations from state transitions rather than raw controls, decoupling action modeling from embodiment-specific hardware and enabling large-scale learning from action-free videos~\cite{dreamdojo}.

\item \textbf{Diffusion \& Flow-Matching Action Head:} Diffusion-based and flow-based action decoders~\cite{diffusion_policy,pi0,gr00tn1_2025} generate actions directly in continuous space, preserving fine-grained dynamics and multimodal behaviors at the cost of higher inference overhead.
\end{itemize}

\subsection{Transition Modeling Paradigms}
\label{sec:wam_paradigms}

\begin{figure}
    \centering
    \includegraphics[width=0.98\linewidth]{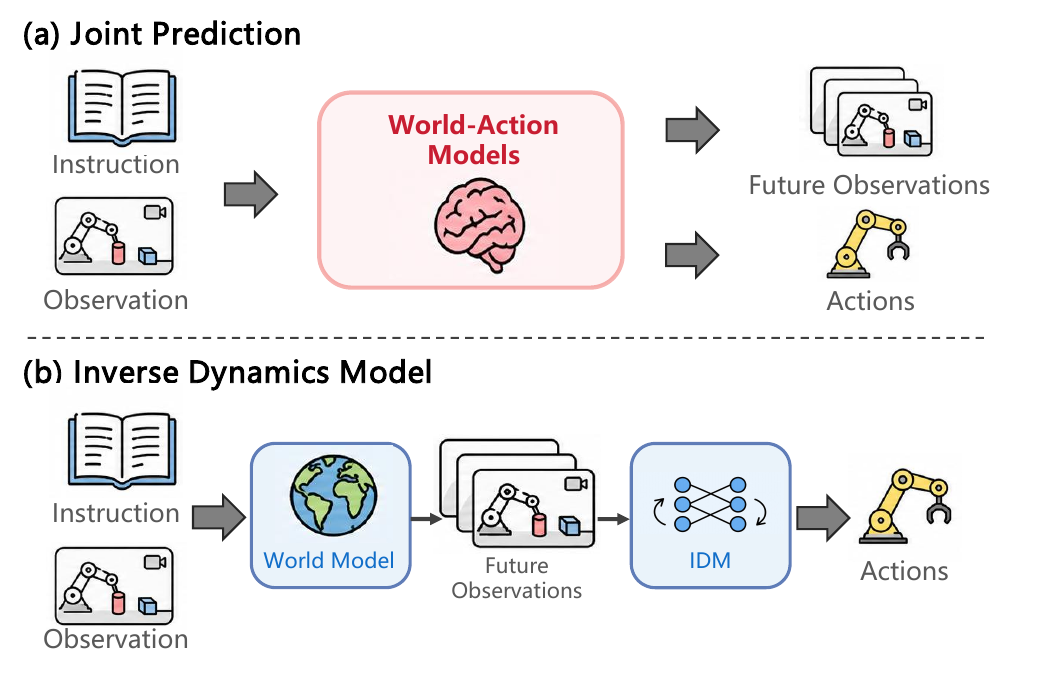}
    \caption{\textbf{Transition Modeling Paradigms.} Joint prediction couples future observation and action generation within a shared process, whereas the inverse dynamics paradigm factorizes this transition by first imagining future states and then inferring executable actions.}
    \label{fig:transition_modeling}
\end{figure}


To instantiate the formulations in~\cref{eq:wam_joint_prediction} and~\cref{eq:wam_plan_than_act}, existing transition modeling of WAMs broadly fall into two complementary paradigms, as illustrated in~\cref{fig:transition_modeling}: \textbf{\emph{(i) joint prediction}}, which couples observation and action generation; and \textbf{\emph{(ii) inverse dynamics}}, which first predicts future states and then infers actions from the resulting transitions.


\subsubsection{Joint Prediction}
For single-step transitions, joint prediction is formulated as
\begin{equation}
\label{eq:joint_prediction}
\pi_{\mathrm{joint}} = p(\mathbf{a}_{t+1}, \mathbf{o}_{t+1} \mid \mathbf{o}_t, \ell),
\end{equation}
which jointly generates the next action and observation conditioned on the current observation $\mathbf{o}_t$ and language instruction $\ell$. The core idea is to unify action generation and future-state prediction under a shared objective, making joint prediction both a direct policy and an auxiliary learning signal that enriches action representations through foresight.

This paradigm has evolved through several architectural stages. Early works such as GR-1~\cite{gr1} and GR-2~\cite{gr2} explored autoregressive co-prediction of actions and future visual representations, demonstrating that language, perception, and control can be modeled within a unified sequence model. With the emergence of diffusion-based generation, later methods adopted continuous denoising formulations for multimodal prediction. VPP~\cite{vpp} combines text-guided video prediction with a diffusion policy head, while UWM~\cite{uwm} unifies video and action diffusion within a coordinated framework.

Recent systems increasingly leverage pretrained video generators as strong temporal priors. WorldVLA~\cite{worldvla} and Motus~\cite{motus} represent two typical designs: the former employs a shared autoregressive token space for text, images, and actions, whereas the latter introduces structured joint attention and separate schedulers for video and action generation. ViPRA~\cite{vipra} and Cosmos Policy~\cite{cosmos_policy} further demonstrate that powerful visual priors can substantially improve joint prediction, suggesting that the central challenge is shifting from jointly modeling actions and observations to effectively aligning pretrained generative dynamics with embodiment-specific control.


\subsubsection{Inverse Dynamics Model}
Under the same single-step assumption, inverse-dynamics WAMs factorize as
\begin{equation}
\label{eq:idm}
\pi_{\mathrm{idm}} = p_{\text{plan}}(\mathbf{o}_{t+1}\mid \mathbf{o}_t,\ell)\cdot p_{\text{idm}}(\mathbf{a}_{t+1}\mid \mathbf{o}_t,\mathbf{o}_{t+1}),
\end{equation}
which first predicts the next observation from the current state and language instruction, then infers the action from the resulting transition. This factorization follows a ``plan-then-act'' paradigm, where future states are generated before actions.

Recent works instantiate this idea with varying degrees of explicitness. In navigation, PathDreamer~\cite{pathdreamer} predicts future visual and semantic states to guide action selection, though planning horizons remain limited. Seer~\cite{seer} incorporates a more structured inverse-dynamics interface by restricting action-token attention to the current and predicted future states. DreamZero~\cite{dreamzero} further adopts a chunk-level autoregressive video-generation framework, although its diffusion-based predict-then-act pipeline still incurs notable inference overhead.

More efficient and implicit IDM variants have recently emerged. LingBot-VA~\cite{lingbotva} combines a Mixture of Transformers (MoT)~\cite{mot} with asynchronous inference to improve modeling capacity and execution efficiency. Fast-WAM~\cite{fastwam} and GigaWorld-Policy~\cite{gigaworld_policy} further decouple visual-generation and action-generation attention mechanisms, enabling joint training while supporting direct action decoding at inference time. These designs are therefore closer to implicit IDMs than fully explicit predict-then-act pipelines. Beyond next-step prediction, goal-reaching dynamics models~\cite{vasan2024revisiting} extend the paradigm to long-horizon planning. For instance, Act2Goal~\cite{act2goal} learns a goal-conditioned IDM that predicts target future states rather than only the immediate next observation.
\(\pi_{0.7}\)~\cite{pi0_7} can be viewed as a goal-conditioned WAM that predicts actions toward a specified goal image, which aligns more naturally with goal-conditioned planning than with next-step joint prediction.

\begin{figure*}
    \centering
    \includegraphics[width=1.\textwidth]{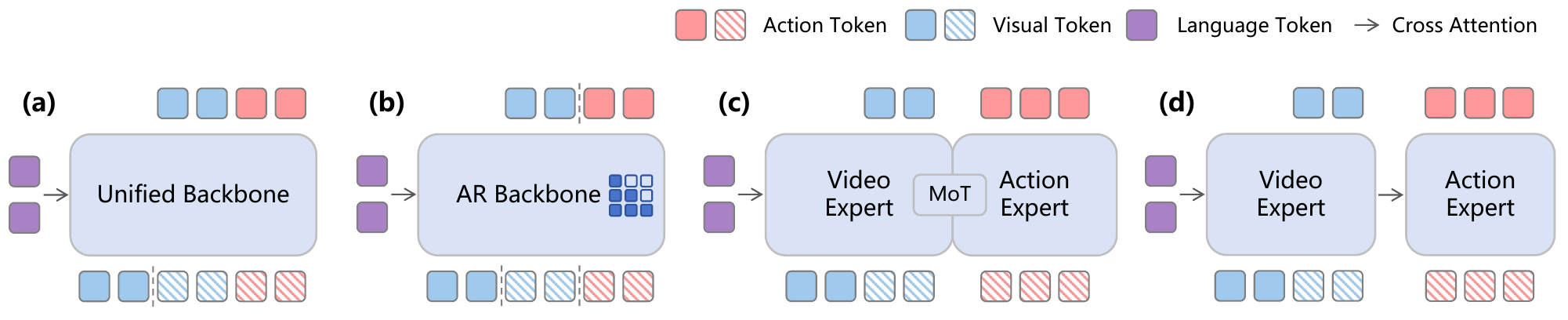}
    \caption{\textbf{Taxonomy of WAM Architectures.} Existing models can be categorized along two orthogonal axes: architectural design and transition modeling paradigm. Structurally, \textbf{end-to-end models} (a, b) process all modalities within a unified backbone, whereas \textbf{dual systems} (c, d) explicitly separate visual and action experts. In terms of transition modeling, \textbf{joint prediction} approaches (a, c) generate future observations and actions simultaneously, while \textbf{inverse dynamics} approaches (b, d) follow a factorized, predict-then-act sequential generation.}
    \label{fig:main_method_model}
\end{figure*}

\subsection{Architecture Structures}
\label{sec:wam_architecture}

To systematically categorize WAM architectures, we examine how multimodal information flows through the network. As shown in~\cref{fig:main_method_model}, existing methods broadly fall into two paradigms: \textbf{\emph{(i) end-to-end systems}} and \textbf{\emph{(ii) dual systems}}.

\subsubsection{End-to-end Systems}
End-to-end WAMs jointly optimize world modeling and action generation within a single backbone. Language, visual observations, embodiment states, and actions are processed through a shared computational pathway, enabling unified representation learning and scalable pre-training. Representative systems such as GR-1~\cite{gr1}, GR-2~\cite{gr2}, and Seer~\cite{seer} map multimodal inputs directly to future visual predictions and control outputs. PAD~\cite{pad} and UWM~\cite{uwm} extend this idea with diffusion transformers that jointly denoise image and action tokens. More recent methods, including WorldVLA~\cite{worldvla}, DreamZero~\cite{dreamzero}, Cosmos Policy~\cite{cosmos_policy}, and ViPRA~\cite{vipra}, further leverage pretrained DiT and video-generation backbones, demonstrating the benefits of large-scale generative pre-training.

A key challenge of end-to-end systems is multimodal alignment across heterogeneous token spaces. Recent works address this through different architectural strategies. Cosmos Policy~\cite{cosmos_policy} transforms non-image signals (e.g., actions, future states, and values) into latent-frame-like representations compatible with video-generation backbones. ViPRA~\cite{vipra} follows a latent action pretraining strategy, learning latent actions from large-scale video~\cite{lapa} before aligning them with robot-specific visual representations. More broadly, end-to-end WAMs increasingly inherit strong pretrained priors from VLMs~\cite{pi0,qwen-vl,gemma_2025} and video diffusion models~\cite{wan}. Under fixed robot-data budgets, richer vision-language and video representations generally improve action generalization and robustness. For example, DWAM~\cite{dwam} reports enhanced robustness and adaptation efficiency, albeit with increased inference cost.

\subsubsection{Dual Systems}
Dual-system WAMs explicitly separate world modeling and action generation into distinct modules connected through an intermediate interface. One module predicts or encodes future world evolution, while the other converts predictive representations into executable actions. This decomposition provides greater modularity and interface-level controllability, as exemplified by VidMan~\cite{vidman} and VPP~\cite{vpp}, although it introduces additional design complexity at the module boundary.

Current dual-system designs are commonly implemented through two interface patterns. The first is MoT-style routing~\cite{mot}, adopted by LingBot-VA~\cite{lingbotva} and Fast-WAM~\cite{fastwam}, where modality-specific projections are fused through shared self-attention in a common latent space. The second is cross-attention transfer~\cite{pi0,mimic_video,gr00tn1_2025}, which explicitly injects visual-language representations into a dedicated action-decoding module. Representative examples in VLAs include $\pi_0$~\cite{pi0}, $\pi_{0.5}$~\cite{pi0_5}, and GR00T-style architectures~\cite{gr00tn1_2025}, where pretrained VLMs first encode multimodal inputs and action experts generate controls via cross-attention conditioning. Mimic-Video~\cite{mimic_video} similarly transfers video features to an action decoder through direct cross-attention. Related systems such as VidMan~\cite{vidman} and VPP~\cite{vpp} also follow this modular design by decoupling predictive video modeling from downstream action generation. 
ImageWAM~\cite{ImageWAM} further adapts this modular paradigm by replacing the video-generation backbone with an image-editing backbone and transferring intermediate representations through a key-value (KV) cache.
Additionally, Helix~\cite{figure_helix} demonstrates the ability to reduce the computational burden of VLMs through a fast-slow dual-system architecture, particularly for humanoid robots.
Overall, the end-to-end versus dual-system taxonomy provides a useful architectural perspective: the former emphasizes unified representation learning, whereas the latter prioritizes modular adaptation and interface design.

\subsection{Policy Training Pipeline}
\label{sec:wam_training}

\begin{figure*}
    \centering
    \includegraphics[width=1.0\linewidth]{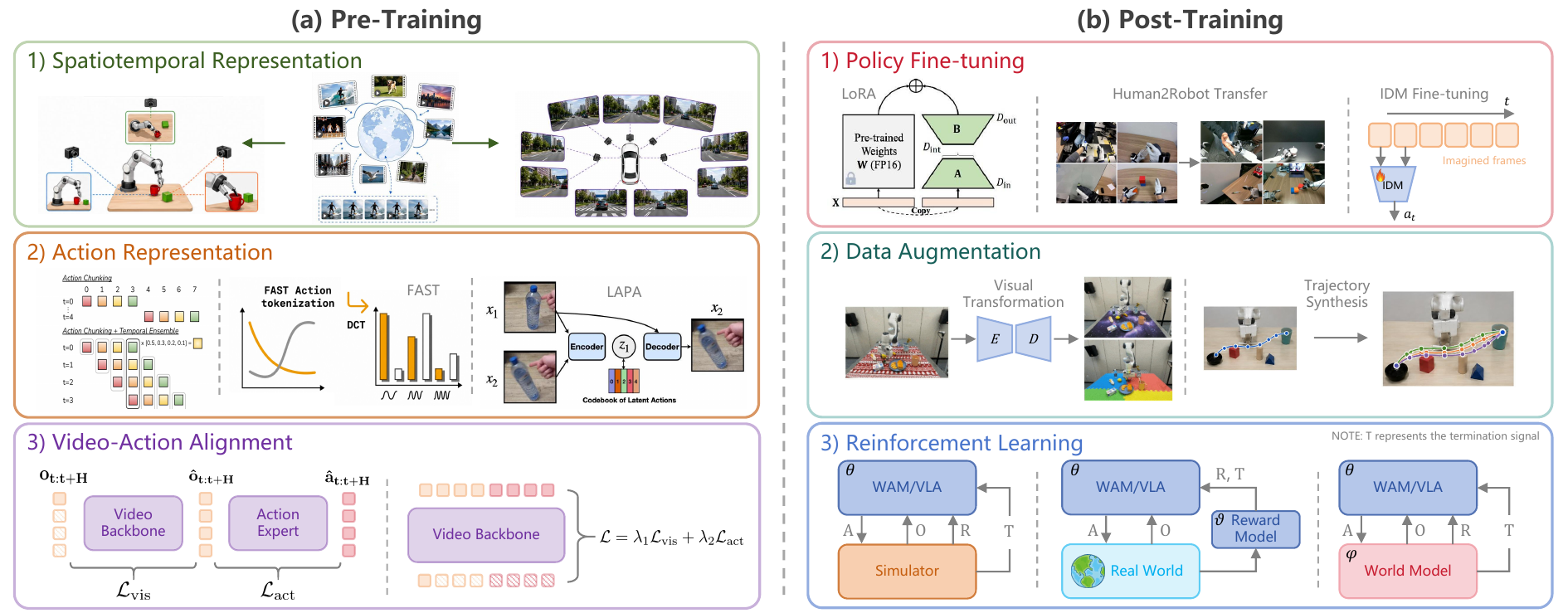}
    \caption{Policy training pipeline for WAMs. (a) Pre-training leverages internet-scale video data to learn spatiotemporal and action representations and align video with actions; (b) post-training adapts the pretrained model through policy fine-tuning, data augmentation, and reinforcement learning. Images adapted from~\cite{fast,lapa,dreamzero}.}
    \label{fig:wam_pipeline}
\end{figure*}

As shown in~\cref{fig:wam_pipeline}, WAM training generally follows a two-stage pipeline: \textbf{pre-training} and \textbf{post-training}. Pre-training learns spatiotemporal and action representations from large-scale video data, while post-training adapts the model through fine-tuning, data augmentation, and reinforcement learning.

\subsubsection{Pre-Training}

The pre-training stage of WAMs centers on spatiotemporal representation learning, action representation learning, and video-action alignment.

\textbf{Spatiotemporal Representation:}
Recent progress in WAMs has been largely driven by pretrained video backbones such as Seedance~\cite{seedance}, Wan~\cite{wan}, Cosmos~\cite{cosmos}, and Sora~\cite{sora}. Trained on large-scale video corpora, these models learn general visual dynamics, object interactions, and action-relevant scene changes, providing richer predictive priors than task-specific policies and simplifying downstream action learning~\cite{dwam}. Multi-view pre-training further improves geometric consistency by preserving 3D structure, object correspondence, and interaction details, benefiting both robot manipulation~\cite{cosmos_policy} and autonomous driving~\cite{omviedrive}.
Beyond pixel-level reconstruction, several works, such as JEPA-WAM~\cite{JEPAWAM} and LaWAM~\cite{LaWAM}, incorporate representation prediction into the WAM framework, enabling dynamics modeling directly in the learned feature space.


\textbf{Action Representation:}
Action representation learning transforms raw control signals into compact, model-friendly formats. One line of work focuses on explicit tokenization or parameterization: OpenVLA~\cite{openvla} uses discrete action tokens with autoregressive decoding, FAST~\cite{fast} applies discrete cosine transforms to decompose actions into high- and low-frequency components, and OpenVLA-OFT~\cite{openvla_oft} adopts continuous representations with bidirectional attention, shifting from next-step prediction toward trajectory-level control.
Another line explores latent action pretraining from videos or action-free trajectories. LAPA~\cite{lapa} learns latent actions from visual changes, AdaWorld~\cite{adaworld} conditions world models on self-supervised latent actions, and ViPRA~\cite{vipra} aligns video-derived latent actions with robot controls during adaptation. VLA-JEPA~\cite{vla_jepa} further combines latent action learning with flow-matching action heads to bridge generic motion representations and embodiment-specific control.


\textbf{Video-Action Alignment:}
A central objective of WAM pre-training is to align video and action representations. UWM~\cite{uwm} identifies four alignment routes: forward dynamics, inverse dynamics, action generation, and video generation. Most systems combine multiple routes. For instance, DreamZero~\cite{dreamzero} couples video generation with inverse dynamics, Fast-WAM~\cite{fastwam} and GigaWorld-Policy~\cite{gigaworld_policy} combine video and action generation, and Cosmos Policy~\cite{cosmos_policy} jointly optimizes action generation, video generation, and forward dynamics.
As discussed, neither joint prediction nor IDM is universally superior~\cite{fastwam}; they reflect different trade-offs. The main challenge is ensuring video representations are both predictive and actionable while maintaining training stability and inference efficiency~\cite{dwam}. Joint-prediction models must balance multimodal generation with accurate action decoding, whereas IDM-based methods rely on sufficiently informative future-state predictions for reliable inverse dynamics.

\subsubsection{Post-Training}
Post-training adapts pretrained WAMs for deployment on specific robot platforms through policy fine-tuning, data augmentation, and reinforcement learning in imagined environments.


\textbf{Policy Fine-tuning:}
Policy fine-tuning adapts pretrained world representations to target sensing modalities, action spaces, and deployment conditions with limited demonstrations or interaction data. Parameter-efficient methods such as LoRA~\cite{lora}, lightweight adapters, and residual action heads enable efficient adaptation of large backbones.
Additional objectives can further improve performance. The image-prediction objective in \cref{eq:wam_vis} enhances generalization to unseen tasks~\cite{dreamzero}, while action-grounding optimization in \cref{eq:wam_inv} supports human-to-robot and robot-to-robot transfer~\cite{vla_jepa}. For IDM-based pipelines, fine-tuning often jointly adapts the planner and inverse-dynamics module to reduce planning mismatch and action-grounding errors~\cite{vidman}. DreamZero~\cite{dreamzero}, for instance, achieves strong cross-embodiment transfer with only about 30 minutes of real-robot data.


\textbf{Data Augmentation:}
Data augmentation improves robustness to visual and environmental variations. One common strategy is label-preserving trajectory rewriting, which re-renders videos with modified appearance, morphology, illumination, or style while preserving task semantics. GigaBrain-0~\cite{gigabrain0}, GigaWorld-0~\cite{gigaworld0}, and Cosmos-Transfer~\cite{cosmos_transfer} use world-model-generated data for real2real, sim2real, and cross-domain transfer.
Another strategy is action-grounded trajectory synthesis. Since video world models generate rollouts without action labels, inverse-dynamics or latent action models recover pseudo-actions to construct synthetic video-action trajectories for policy learning and evaluation, as explored by DreamGen~\cite{dreamgen} and WorldArena~\cite{worldarena}.

\begin{figure*}[t]
    \centering
    \includegraphics[width=0.9\textwidth]{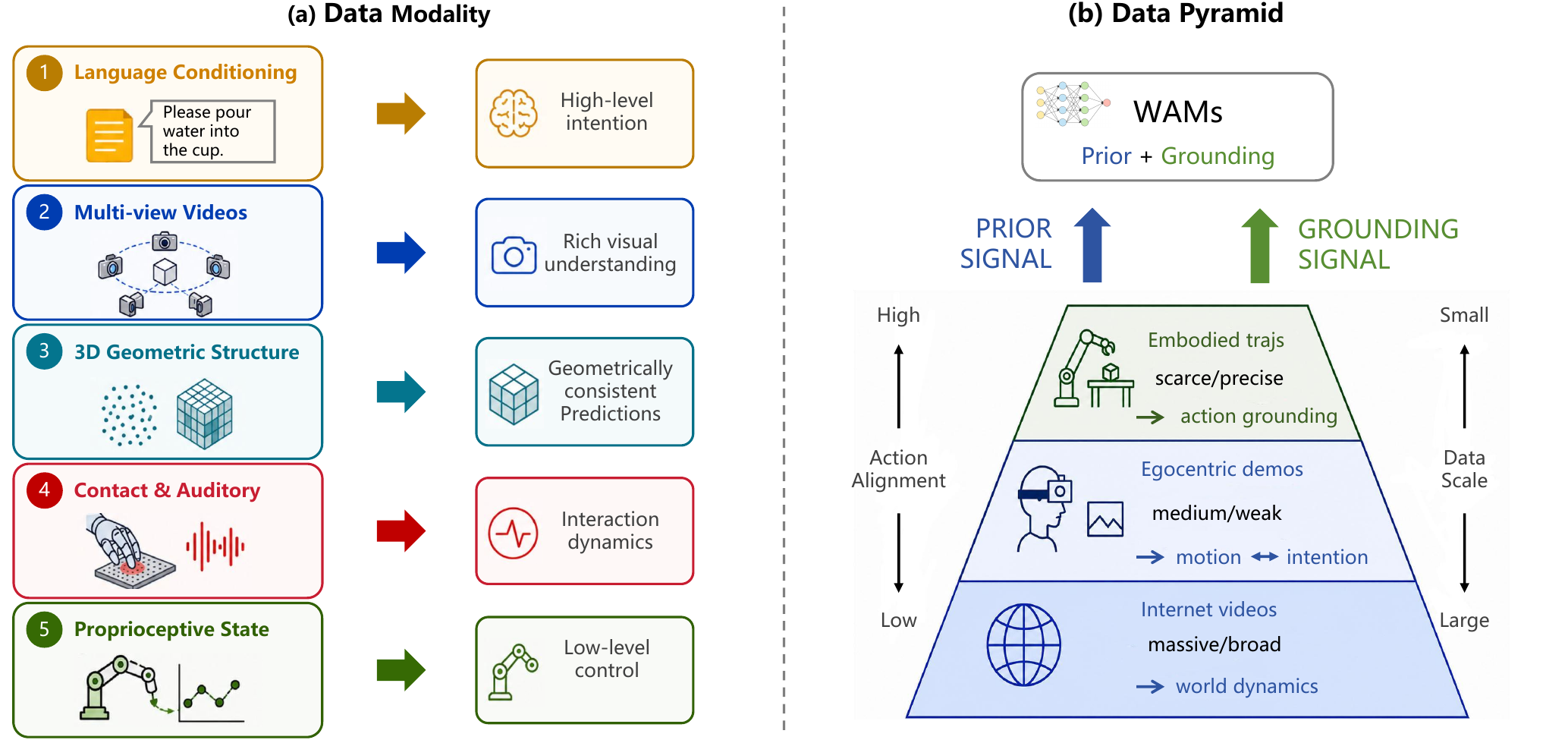}
    \caption{Data modality and scaling structure for \wams{}. The left panel illustrates how different sensing modalities expose complementary action-relevant state variables, while the right panel shows a data-pyramid view in which large action-free video corpora provide broad world priors and smaller embodied trajectories provide action grounding.}
    \label{fig:wam_data_modality_scaling}
\end{figure*}

\textbf{Reinforcement Learning:}
In large language models, reinforcement-learning-based post-training methods such as RLHF~\cite{christiano2017deep,stiennon2020learning} and OPD~\cite{OPD} have been widely adopted. 
In robotics, RL post-training has likewise demonstrated substantial benefits for dexterous and contact-rich manipulation~\cite{pi0_6,gigabrain0_5,rl_token}.
Embodied RL generally follows three routes: simulation-based RL exploits privileged states and dense rewards but faces the sim2real gap; real-world RL, such as SERL~\cite{serl} and RL-token~\cite{rl_token}, avoids simulation transfer but incurs substantial interaction and safety costs; and world-model-based RL~\cite{world4rl,world_vla_loop,WAMOPD} optimizes policies through imagined rollouts, reducing physical interaction at the risk of model bias.
A typical world model-based pipeline trains an action-conditioned world model, a reward predictor~\cite{MVWAM}, and an imitation-initialized policy before refinement through imagined trajectories~\cite{wmpo}. World4RL~\cite{world4rl} performs PPO optimization with a diffusion world model, while World-VLA-Loop~\cite{world_vla_loop} connects imagined experience with real-world improvement. WAM-OPD~\cite{WAMOPD} instead provides dense teacher supervision on student-induced trajectories. Although Cosmos Policy~\cite{cosmos_policy} jointly predicts future observations, values, and actions, their effective integration for scalable WAM post-training remains an open problem.

\subsection{Data Modality, Scaling, and Mixture}
\label{sec:data_modality_scaling}


Data modality and scale are critical to both VLAs and WAMs, as illustrated in~\cref{fig:wam_data_modality_scaling}. 
The choice of modality determines what information the model observes and how each modality
contributes to decision-making, while scale influences the richness of learned dynamics and the availability of action supervision.

\subsubsection{Data Modality}

The following summarizes major data modalities and their implications for WAM design.

\textbf{Language Conditioning:} Language provides task-level conditioning. In VLAs, it is often used as contextual input or for subtask planning~\cite{pi0_5}, whereas WAMs typically encode language separately and fuse it with visual or state features through cross-attention.

\textbf{Multi-view Video:} Visual observations provide rich environmental cues for decision-making and control. Multi-view inputs further improve spatial understanding, 
although generating consistent multi-view predictions remains challenging for WAMs~\cite{MultiviewVideo}.

\textbf{3D Geometric Structure:} 3D representations offer unified spatial structures and explicit geometric constraints~\cite{particleformer,manigauss_pp}. While prediction in occupancy or geometric spaces has shown promise~\cite{occworld,drive_occworld,gwm}.
4DGSWAM~\cite{4DGSWAM} and 4DWAM\cite{4DWAM} explore general 3D world prediction, although this research direction remains at an early stage.

\textbf{Contact and Auditory Feedback:} Tactile~\cite{HiTacWAM,FAWAM} and auditory signals reveal interaction events that may be difficult to infer from vision alone, especially in industrial applications~\cite{microsoft_advancing_ai}. Existing work either incorporates these sensors directly~\cite{vtla,audio_wm} or predicts them from visual inputs~\cite{vt_wm}, suggesting their potential for interaction-aware world modeling.

\textbf{Proprioception State:} Robot states such as joint positions and end-effector poses provide direct physical information, complementing visual observations under occlusion and supporting action generation.
Recent works~\cite{cosmos_policy} further use predicted actions as part of the state to reduce OOD issues and enhance temporal consistency.

\subsubsection{Data Scaling and Mixture}

As illustrated in~\cref{fig:wam_data_modality_scaling}, different data sources provide complementary supervision for WAMs. We categorize them into three groups: task-agnostic internet videos, task-relevant egocentric demonstrations~\cite{egodex,ego4d}, and embodiment-specific trajectories~\cite{oxe}. The first two mainly provide visual dynamics priors, while the last supplies direct action supervision.

\textbf{Internet Videos:}
Internet-scale videos primarily support temporal prediction and representation learning. GR-1~\cite{gr1} and GR-2~\cite{gr2} use such data for visual pre-training, while LAPA~\cite{lapa} and DreamDojo~\cite{dreamdojo} infer latent actions from unlabeled videos. Large video generators such as Wan~\cite{wan} and Cosmos~\cite{cosmos} further demonstrate the value of web-scale video corpora for learning world dynamics.

\textbf{Egocentric Demonstrations:}
UMI~\cite{UMI} expands manipulation data through robot-free, handheld demonstrations transferable to robot policies. Egocentric datasets~\cite{egodex,ego4d} provide rich first-person interaction videos, with actions recovered through instrumented gloves or vision-based hand-pose estimation~\cite{activeglasses}. EgoScale~\cite{egoscale} further highlights the potential of human demonstrations for cross-embodiment transfer.
Dyna-2~\cite{robotics2026dyna} scales first-person human manipulation videos to one million hours, demonstrating the advantage of WAMs over VLAs in learning from videos without hand-action labels.

\textbf{Embodied Trajectories:}
Embodied trajectories ground WAMs in executable control by coupling observations with robot states, actions, and physical outcomes. Although much smaller than web-scale video datasets, resources such as Open X-Embodiment~\cite{oxe}, DROID~\cite{droid}, and BridgeData~\cite{bridgev2} provide essential action-conditioned supervision across tasks, embodiments, and contact-rich interactions, serving as the key bridge between broad visual priors and controllable world prediction. LBM~\cite{LBM} performs large-scale pre-training on these real robot data using diffusion policy~\cite{diffusion_policy} and demonstrates robustness under distribution shift.

%% file: final_chapters/Sec-IV.tex
\input{final_chapters/tables/table_application.tex}

\section{Applications}
\label{sec:applications}

WAMs are used differently across embodied domains. Their value may lie in representation learning before policy training, in imagined look-ahead during inference, or in synthetic trajectories used for policy improvement. 
We therefore organize this section by application domain, including robot manipulation (\cref{sec:manipulation}), navigation (\cref{sec:navigation}), autonomous driving (\cref{sec:driving}), and examine the role that prediction plays in each domain.

\subsection{Robot Manipulation}
\label{sec:manipulation}
Manipulation is the most common testbed for WAMs because it couples visual prediction with physically constrained action. 
A useful rollout must accurately model object states, contact dynamics, occlusions, and task progress rather than merely appear visually plausible.
Existing work falls into three broad groups: video-action policies, structured state models, and imagination-based policy optimization.

\subsubsection{Video-Action and Unified Policies}
Video-action models use predictive pre-training to shape action representations. GR-1~\cite{gr1} and GR-2~\cite{gr2} establish a representative recipe: pretrain a temporal video backbone and adapt it to demonstrations so that future observations and robot actions share a trajectory-level representation. Diffusion and flow-based variants such as PAD~\cite{pad}, UWM~\cite{uwm}, VPP~\cite{vpp}, Seer~\cite{seer}, VidMan~\cite{vidman}, Cosmos Policy~\cite{cosmos_policy}, and ViPRA~\cite{vipra} further tighten this connection through denoising, latent action alignment, or lightweight control adapters.
Unified WAMs make the prediction-control interface explicit at inference time. WorldVLA~\cite{worldvla}, Motus~\cite{motus}, and LingBot-VA~\cite{lingbotva} jointly generate observations and actions, while World2Act~\cite{world2act} and Act2Goal~\cite{act2goal} use predicted future states or goals as intermediate control targets. In contrast, VLA-JEPA~\cite{vla_jepa}, LAWM~\cite{lawm}, and HiLAM~\cite{hilam} mainly use predictive objectives to improve latent representations, avoiding costly online video rollout during deployment.

\subsubsection{Structured State Models}
RGB rollouts are often underspecified for contact-rich manipulation. Geometry- and physics-oriented WAMs therefore predict representations closer to control variables: ManiGaussian++~\cite{manigauss_pp} and GWM~\cite{gwm} model dynamic scenes with Gaussian representations, ParticleFormer~\cite{particleformer} captures deformable and multi-material interactions, PointWorld~\cite{pointworld} scales point-cloud dynamics, and PIN-WM~\cite{pin_wm} adds physics-informed constraints. Multimodal models such as VT-WM~\cite{vt_wm}, Cloth-WM~\cite{cloth_wm}, and Audio-WM~\cite{audio_wm} further capture contact, compliance, or event cues that are weakly visible in RGB.

\subsubsection{Imagination-based Policies}
A complementary direction treats the world model as an imagined environment. World4RL~\cite{world4rl}, DreamPlan~\cite{dreamplan}, VAMPO~\cite{vampo}, PRWM~\cite{prwm}, WMPO~\cite{wmpo}, EVA~\cite{eva_vwm}, and RISE~\cite{rise} optimize or evaluate policies through imagined interaction. VLA-MBPO~\cite{vla_mbpo}, GPC~\cite{gpc}, RoboHorizon~\cite{robohorizon}, World-VLA-Loop~\cite{world_vla_loop}, and IWS~\cite{iws} further close the loop between policy failures, world-model rollouts, and synthetic trajectory generation. Across these lines, the core question is whether predicted futures improve executable control 
in contact-rich settings, under distribution shift, and over long horizons.

\subsection{Navigation}
\label{sec:navigation}
Navigation is less contact-sensitive than manipulation, but it relies more heavily on long-horizon spatial reasoning under partial observability, using egocentric visual context to make sequential waypoint decisions.
WAMs are therefore used mainly as look-ahead modules: they predict future egocentric views, semantic maps, or latent states that support waypoint or action selection.

\subsubsection{Look-Ahead Planning}
Pathdreamer~\cite{pathdreamer} predicts future panoramas and semantic maps for waypoint selection, while DreamWalker~\cite{dreamwalker} combines latent imagination with tree search to avoid long pixel-space rollouts. NWM~\cite{nwm} and MonoDream~\cite{monodream} extend this paradigm across navigation domains and monocular panoramic settings. 
Platform-aware future poses and failure probabilities can be predicted from terrain geometry, proprioceptive history, and candidate velocity commands to support MPPI-based navigation~\cite{RothP-RSS-25}.
These methods keep planning modular: the world model supplies candidate futures, and a downstream planner selects actions.


\subsubsection{Latent Planning and WM-RL}
Other systems reduce the predictive interface to latent plans or imagined training environments. X-MOBILITY~\cite{x_mobility} trains a mobile policy with reinforcement learning inside a latent world model, whereas InternVLA-N1~\cite{internvla_n1} learns compact latent plans from unlabeled navigation trajectories and translates them into real-time commands. WMP~\cite{lai2025world} uses the recurrent latent state of a world model trained on depth images and proprioception to guide visual legged locomotion. RWM~\cite{li2025robotic} serves as a learned neural simulator for optimizing robotic control policies over long autoregressive imagined rollouts. Thus, these navigation WAMs need not generate explicit images; their practical value is measured by improved waypoint selection, long-horizon consistency, and robustness under partial observability.

\subsection{Autonomous Driving}
\label{sec:driving}

Driving WAMs must satisfy real-time, safety, and multi-agent constraints. Prediction is used not only for control, but also for scenario synthesis and structured scene evolution. This leads to three recurring roles: controllable video generation, occupancy-based forecasting, and unified perception-prediction-planning.

\subsubsection{Generation and Structured Scene Prediction}
Generative driving WAMs synthesize future traffic scenes conditioned on commands, trajectories, or text. DriveDreamer~\cite{drivedreamer}, GAIA-1~\cite{gaia_1}, DriveDreamer-2~\cite{drivedreamer_2}, GAIA-2~\cite{gaia_2}, DrivingWorld~\cite{drivingworld}, and DrivingGen~\cite{drivinggen} use this capability for data augmentation, scenario coverage, or policy-conditioned evaluation. DriveVLA-W0~\cite{drivevla_w0} further studies how such generated data affects downstream driving policies.
Occupancy-based WAMs replace pixel futures with planning-oriented 3D states. OccWorld~\cite{occworld} and Driving in the Occupancy World~\cite{drive_occworld} forecast semantic occupancy under action or temporal conditioning, aligning prediction with free space, dynamic agents, and collision-related evaluation.

\subsubsection{Unified Driving Agents and WM-RL}
Unified driving WAMs couple scene prediction and action generation. ADriver-I~\cite{adriver_i}, DrivingGPT~\cite{drivinggpt}, Doe-1~\cite{doe_1}, and OccLLaMA~\cite{occllama} integrate future observation, trajectory, occupancy, or language-conditioned action modeling. ProphetDWM~\cite{prophetdwm}, VaViM/VaVAM~\cite{vavim_vavam}, DriveLaW~\cite{drivelaw}, and AutoMoT~\cite{automot} further emphasize long-horizon consistency, structured latents, and efficient control.
Driving also motivates world-model RL. DreamerAD~\cite{dreamerad} and World4Drive~\cite{world4drive} train policies through imagined rollouts, which is attractive for rare and safety-critical events but depends strongly on action-conditioned dynamics fidelity. The main evaluation challenge is to connect visual realism, traffic consistency, safety, and closed-loop driving performance.

%% file: final_chapters/tables/table_application.tex
%
%

\definecolor{vamblue}{HTML}{1F77B4}
\definecolor{uniteal}{HTML}{17A2A2}
\definecolor{pwpurple}{HTML}{8E44AD}
\definecolor{rlorange}{HTML}{D17A22}
\definecolor{threedbrown}{HTML}{8B5A2B}
\providecommand{\PVAM}{\colorbox{vamblue!15}{\textcolor{vamblue}{\textbf{{VAM}}}}}
\providecommand{\PUni}{\colorbox{uniteal!15}{\textcolor{uniteal}{\textbf{{Uni}}}}}
\providecommand{\PPW}{\colorbox{pwpurple!15}{\textcolor{pwpurple}{\textbf{{WM-Plan}}}}}
\providecommand{\PRL}{\colorbox{rlorange!15}{\textcolor{rlorange}{\textbf{{RL}}}}}
\providecommand{\PThreeD}{\colorbox{threedbrown!15}{\textcolor{threedbrown}{\textbf{{3D}}}}}
\providecommand{\WAMCapChip}[3][15]{%
  \begingroup
  \setlength{\fboxsep}{0.5pt}%
  \colorbox{#2!#1}{\textcolor{#2}{\textbf{#3}}}%
  \endgroup
}

\begin{table*}[!ht]
\centering
\scriptsize
\caption{Unified summary of WAMs in~\cref{sec:applications}, grouped by application domain. Category labels: \WAMCapChip{vamblue}{VAM} Video-Action Models, \WAMCapChip{uniteal}{Uni} Unified WAMs (joint observation and action generation), \WAMCapChip{pwpurple}{WM-Plan} WMs-guided Planning (predictions from WMs and action generated by external planner), \WAMCapChip{rlorange}{RL} WM-RL (imagination-based RL), \WAMCapChip{threedbrown}{3D} 3D/Geometry-based.}
\label{tab:application-methods}
\setlength{\tabcolsep}{3pt}
\setlength{\fboxsep}{1pt}
\setlength{\extrarowheight}{0pt}
\renewcommand{\arraystretch}{0.5}
\begin{tabular*}{\textwidth}{@{\extracolsep{\fill}}
  l c l l
  l c l l
  @{}}
\toprule
\textbf{Method} & \textbf{Paradigm} & \textbf{Backbone} & \textbf{Eval / Note} &
\textbf{Method} & \textbf{Paradigm} & \textbf{Backbone} & \textbf{Eval / Note} \\
\midrule
\rowcolor{gray!15}
\multicolumn{8}{@{}l}{\emph{\textbf{Manipulation}}} \\
GR-1~\cite{gr1}              & \PVAM    & GPT       & LIBERO, CALVIN
  & EVA~\cite{eva_vwm}         & \PRL    & video WM  & RoboTwin 2.0 \\
GR-2~\cite{gr2}              & \PVAM    & GPT, 38M  & LIBERO, CALVIN
  & RISE~\cite{rise}           & \PRL    & latent    & contact-rich \\
PAD~\cite{pad}               & \PUni    & DiT       & LIBERO
  & VLA-MBPO~\cite{vla_mbpo}   & \PRL    & video WM  & LIBERO-Plus \\
UWM~\cite{uwm}               & \PUni    & 4-mode DiT & LIBERO
  & GPC~\cite{gpc}             & \PRL    & rollout WM & Push-T \\
VPP~\cite{vpp}               & \PVAM    & video pred. & dexterous
  & WMPO~\cite{wmpo}           & \PRL    & pixel WM  & LIBERO \\
Seer~\cite{seer}             & \PVAM    & GPT+mask  & 100K demos
  & RoboHorizon~\cite{robohorizon} & \PRL & LLM+WM   & sparse reward \\
VidMan~\cite{vidman}         & \PVAM    & vid.\ diff.+adapt. & Bridge
  & World-VLA-Loop~\cite{world_vla_loop} & \PRL & feedback & ManiSkill \\
Cosmos Policy~\cite{cosmos_policy} & \PVAM & Cosmos-P2 & bimanual
  & IWS~\cite{iws}             & \PRL    & video WM  & 10-min @15FPS \\
ViPRA~\cite{vipra}           & \PVAM    & flow-match & 100-200 demos
  & ManiGaussian++~\cite{manigauss_pp} & \PThreeD & Gaussian & bimanual \\
GigaWorld-Policy~\cite{gigaworld_policy} & \PUni & action-cent. & DROID
  & ParticleFormer~\cite{particleformer} & \PThreeD & particles & deformable \\
Fast-WAM~\cite{fastwam}      & \PVAM    & video pred. & LIBERO, RoboTwin
  & GWM~\cite{gwm}             & \PThreeD & VAE+diff & MetaWorld \\
WorldVLA~\cite{worldvla}     & \PUni    & Chameleon AR & LIBERO
  & PointWorld~\cite{pointworld} & \PThreeD & point cloud & 2M+ traj \\
Motus~\cite{motus}           & \PUni    & MoT       & RoboTwin
  & PIN-WM~\cite{pin_wm}       & \PThreeD & physics-inf. & non-prehensile \\
LingBot-VA~\cite{lingbotva}  & \PUni    & causal-diff MoT & AgiBot
  & VT-WM~\cite{vt_wm}         & \PThreeD & visuo-tactile & contact \\
World2Act~\cite{world2act}   & \PUni    & LLM+skill & RoboCasa
  & Cloth-WM~\cite{cloth_wm}   & \PThreeD & DreamerV2 & cloth \\
Act2Goal~\cite{act2goal}     & \PVAM    & goal-cond. & RoboTwin 2.0
  & Audio-WM~\cite{audio_wm}   & \PThreeD & acoustic  & liquid pour \\
VLA-JEPA~\cite{vla_jepa}     & \PVAM    & JEPA+flow & LIBERO
  & World4RL~\cite{world4rl}   & \PRL    & diff.\ WM & frozen env \\
LAWM~\cite{lawm}             & \PVAM    & latent action & LIBERO
  & DreamPlan~\cite{dreamplan} & \PRL    & video WM  & VLA finetune \\
HiLAM~\cite{hilam}           & \PVAM    & hier.\ latent & long-horizon
  & VAMPO~\cite{vampo}         & \PRL    & GRPO/denoise & LIBERO, CALVIN \\
PRWM~\cite{prwm}             & \PRL    & contrastive RL & DROID
  &                            &         &           &  \\
\midrule
\rowcolor{gray!15}
\multicolumn{8}{@{}l}{\emph{\textbf{Navigation}}} \\
Pathdreamer~\cite{pathdreamer} & \PPW   & panoramic gen. & R2R
  & MonoDream~\cite{monodream}  & \PPW   & monocular & R2R-CE, RxR-CE \\
DreamWalker~\cite{dreamwalker} & \PPW   & latent+MCTS & R2R, VLN-CE
  & X-MOBILITY~\cite{x_mobility} & \PRL  & latent WM env & mobile robot \\
NWM~\cite{nwm}                & \PPW   & multi-domain & Habitat, Ego4D
  & InternVLA-N1~\cite{internvla_n1} & \PVAM & dual-system latent & R2R-CE \\
  perceptive FDM~\cite{RothP-RSS-25} &  \PPW & GRU & ANYmal, quadruped & RWM~\cite{li2025robotic} & \PRL & GRU& A1, quadruped \\
  WMP~\cite{lai2025world} & \PRL & RSSM & Isaac & & & &\\
\midrule
\rowcolor{gray!15}
\multicolumn{8}{@{}l}{\emph{\textbf{Autonomous Driving}}} \\
DriveDreamer~\cite{drivedreamer}     & \PPW & video gen. & nuScenes
  & ADriver-I~\cite{adriver_i}        & \PUni & AR img+act & nuScenes \\
GAIA-1~\cite{gaia_1}                 & \PPW & text-video & large-scale
  & DrivingGPT~\cite{drivinggpt}      & \PUni & large AR video & NAVSIM \\
DriveDreamer-2~\cite{drivedreamer_2} & \PPW & multi-cam.\ gen. & nuScenes
  & Doe-1~\cite{doe_1}                & \PUni & autoregressive & nuScenes \\
GAIA-2~\cite{gaia_2}                 & \PPW & multi-cam.\ scaled & OpenDV
  & OccLLaMA~\cite{occllama}          & \PUni & occ.+language & nuScenes-QA \\
DrivingWorld~\cite{drivingworld}     & \PPW & long-horizon AR & nuScenes
  & ProphetDWM~\cite{prophetdwm}      & \PUni & long-horizon & NAVSIM, nuPlan \\
DrivingGen~\cite{drivinggen}         & \PPW & policy-cont.\ eval & nuScenes
  & VaViM/VaVAM~\cite{vavim_vavam}    & \PUni & video AR+action AR & OpenDV \\
OccWorld~\cite{occworld}             & \PPW & voxel tokens & nuScenes-Occ
  & DriveLaW~\cite{drivelaw}          & \PUni & language-action & nuScenes \\
Drive-OccWorld~\cite{drive_occworld} & \PPW & multi-step occ. & nuScenes-Occ
  & AutoMoT~\cite{automot}            & \PUni & async.\ MoT & Bench2Drive \\
DreamerAD~\cite{dreamerad}           & \PRL & Dreamer-AD & nuScenes, nuPlan
  & World4Drive~\cite{world4drive}    & \PRL & virtual env RL & NAVSIM, nuPlan \\
\bottomrule
\end{tabular*}
\end{table*}

%% file: final_chapters/Sec-V.tex
\definecolor{simblue}{HTML}{2E6FB7}
\definecolor{realgreen}{HTML}{2E8B57}
\definecolor{mixedpurple}{HTML}{8E44AD}
\providecommand{\Sim}{\colorbox{simblue!18}{\strut\textcolor{simblue}{\textbf{Sim}}}}
\providecommand{\Real}{\colorbox{realgreen!18}{\strut\textcolor{realgreen}{\textbf{Real}}}}
\providecommand{\Mixed}{\colorbox{mixedpurple!18}{\strut\textcolor{mixedpurple}{\textbf{Mixed}}}}

\section{Datasets, Benchmarks, and Evaluation Metrics}
\label{sec:datasets}
Evaluation resources for WAMs have different purposes. Datasets provide offline demonstrations, trajectories, or video-language pairs for pre-training and adaptation; benchmarks define task protocols and metrics; simulators provide the physical or visual environment in which closed-loop policies are tested. Since many resources are shared across application domains, we organize this section by resource type. 

\subsection{Metric Families and Evaluation Protocols}
\label{sec:metrics-families}

\input{final_chapters/tables/table_metric_glossary.tex}

\cref{tab:metric-glossary} expands the acronyms and gives compact definitions for the metric families used for WAMs.


Task-level metrics such as SR, route completion, driving score, episode return, and chain length are closest to the final control objective, but they rarely isolate whether success comes from better prediction, stronger imitation priors, or task-specific policy learning~\cite{libero,calvin,navsim,atari}.
Trajectory metrics such as ADE/FDE, L2 displacement, NE, SPL, nDTW, ATE, and RPE measure path or pose consistency and are useful for planning-heavy WAMs, but low trajectory error does not guarantee safety under feedback. Generation metrics such as FID, FVD, LPIPS, PSNR, and SSIM evaluate visual fidelity; they are necessary for video-based WAMs but insufficient for action grounding. Semantic, language, and occupancy metrics such as mIoU, mAP, NDS, BLEU, CIDEr, and ROUGE assess structured outputs, while latency, collision, comfort, and infraction metrics capture deployment constraints. WAM evaluations are therefore most informative when it reports both predictive quality and downstream control utility.

\input{final_chapters/tables/tab_unified_benchmark_v3}

\subsection{Robot Demonstration and Human Video Corpora}
\label{sec:rc}

Paired robot resources primarily support pre-training and embodiment adaptation rather than serving as standalone evaluation benchmarks. As summarized in \cref{tab:unified-benchmarks}, corpora such as DROID, BridgeData V2, Open X-Embodiment, and AgiBot-World provide action-grounded demonstrations at different scales across diverse embodiments~\cite{droid,bridgev2,oxe,agibotworld}. Their utility for WAMs is typically assessed using downstream task success, instruction-following accuracy, and rollout quality~\cite{gr1,gr2,uwm,vidman,cosmos_policy}.
A second group comprises human video resources spanning egocentric recordings and web-scale collections, including Something-Something v2 (SSv2), Ego4D, EgoDex, Kinetics-700, and HowTo100M. These datasets expand motion, interaction, and language priors far beyond the scale of paired robot data, but they do not directly specify robot actions~\cite{ssv2,ego4d,egodex,kinetics,howto100m}. Consequently, their contribution is usually assessed after latent action learning, inverse-dynamics alignment, or robot-data fine-tuning~\cite{lapa,vipra,dreamzero,gr2}. The distinction reflected in \cref{tab:unified-benchmarks} is therefore important: unpaired human video improves predictive priors, whereas paired robot trajectories provide the action grounding required for control.

\subsection{Simulated Manipulation Task Suites}
\label{sec:sm}

Simulated manipulation suites provide controlled closed-loop evaluation for WAMs. The resources summarized in \cref{tab:unified-benchmarks} span language-conditioned long-horizon tasks, multi-task manipulation, dual-arm control, kitchen-scale scene generalization, synthetic demonstration generation, and focused contact-rich probes~\cite{libero,calvin,metaworld,robotwin,robocasa,mimicgen,rlbench,maniskill,simplerenv,diffusion_policy}. This diversity matters because a WAM that performs well on short-horizon single-object manipulation may still fail under instruction chaining, contact ambiguity, or scene-level generalization.
The main evaluation signal in these suites is task success, but the metric profile varies with the benchmark's purpose. LIBERO, MetaWorld, RoboCasa, RLBench, ManiSkill, SimplerEnv, and Push-T emphasize success rate, normalized score, or contact/task-specific measures, whereas CALVIN highlights long-horizon chain completion~\cite{libero,metaworld,robocasa,rlbench,maniskill,simplerenv,diffusion_policy,calvin}. Video-oriented WAM papers additionally report rollout fidelity metrics such as FVD, LPIPS, PSNR, and SSIM~\cite{gr1,gr2,pad,uwm,vpp}. Thus, the simulated manipulation block in \cref{tab:unified-benchmarks} should be read as a spectrum from policy evaluation to prediction-quality evaluation, rather than as a single interchangeable benchmark family.

\subsection{Autonomous Driving Datasets and Simulators}
\label{sec:ad}

Driving resources exhibit a clear separation between open-loop prediction and closed-loop control. Recorded datasets such as nuScenes, Waymo Open Motion, BDD100K, KITTI, Argoverse, ZOD, and OpenDV support visual generation, trajectory prediction, occupancy forecasting, and perception evaluation~\cite{nuscenes,waymo,bdd100k,kitti,argoverse,zod,opendv}. By contrast, nuPlan-derived NAVSIM protocols, CARLA, and Bench2Drive primarily support closed-loop planning and policy assessment~\cite{nuplan,navsim,navsimv2,carla,bench2drive}. \cref{tab:unified-benchmarks} makes this split explicit through its mix of real, simulated, and mixed resources.
For driving WAMs, metric choice strongly determines the interpretation of progress. FID, FVD, and LPIPS assess whether generated traffic videos look realistic~\cite{fid,fvd,lpips}; ADE/FDE, L2, mIoU, mAP, and NDS assess structured prediction~\cite{nuscenes,waymo,bdd100k}; and DS, RC, IS, PDMS, TTC, comfort, and collision-related metrics assess planning utility and safety~\cite{navsim,navsimv2,bench2drive,carla}. A driving WAM should therefore not be judged by video realism alone. The driving block in \cref{tab:unified-benchmarks} illustrates how metrics for video realism, structured scene prediction, and closed-loop safety should be combined.

\subsection{Navigation Environments, Datasets, and Benchmarks}
\label{sec:nav}

Navigation resources usually combine a 3D environment with an instruction, point-goal, image-goal, or real-trajectory protocol. Matterport3D-style scans and Habitat-style simulation provide the spatial substrate, while R2R/R2R-CE, RxR/RxR-CE, and ScaleVLN define instruction-following and data-scaling settings~\cite{matterport3d,habitat,r2r,r2r_ce,rxr,scalevln}. Real-robot or field datasets such as RECON~\cite{recon}, SCAND~\cite{scand}, CoPeD~\cite{Zhou_2024}, and TartanDrive~\cite{tartandrive} complement these simulated protocols by exposing social, off-road, multi agent or platform-specific dynamics. Their reported scales and typical evaluation measures are summarized in \cref{tab:unified-benchmarks}.
For WAMs, navigation evaluation should capture more than successful arrival. SR measures task completion, SPL and path length reflect efficiency, NE and nDTW capture route fidelity, and ATE/RPE assess pose consistency when prediction or mapping is involved~\cite{spl_eval,ndtw,tum_rgbd}. This metric family is particularly relevant for look-ahead WAMs, where predicted egocentric views or latent states may improve waypoint selection even when the model does not generate photorealistic future frames~\cite{pathdreamer,dreamwalker,nwm,monodream}.

\subsection{Foundational RL Benchmarks}
\label{sec:rl}

Foundational RL benchmarks predate recent WAMs for robotics, but they remain useful diagnostic settings for latent dynamics, imagined rollouts, sample efficiency, and generalization. As shown in \cref{tab:unified-benchmarks}, DMControl and Atari-style protocols respectively test continuous and discrete control under compact task suites; Atari 100k emphasizes sample efficiency; Procgen, Craftax, and DMLab stress generalization and long-horizon exploration; and BSuite provides targeted diagnostics for core RL capabilities~\cite{dmcontrol,atari,atari100k,procgen,craftx,dmlab,bsuite}.
These benchmarks are not substitutes for embodied robot evaluation because they abstract away many sensing, contact, embodiment, and safety constraints. Their role is instead methodological: they help establish whether a predictive model supports planning, value learning, or policy improvement before the same ideas are transferred to manipulation, navigation, or driving~\cite{dreamer,dreamerv2,dreamerv3,td_mpc,td_mpc2}. FMB is an important bridge case because it reports real-world manipulation success and cycle time, linking RL-style evaluation back to physical execution~\cite{fmb}.

%% file: final_chapters/tables/table_metric_glossary.tex
\begin{table*}[!t]
\centering
\scriptsize
\begin{threeparttable}
\caption{Compact definitions of metrics used for WAMs. 
}

\label{tab:metric-glossary}

\setlength{\tabcolsep}{3pt}
\setlength{\extrarowheight}{0pt}
\renewcommand{\arraystretch}{1.0}

\begin{tabular*}{\textwidth}{
@{\extracolsep{\fill}}
>{\raggedright\arraybackslash}p{0.36\textwidth}
>{\raggedright\arraybackslash}p{0.62\textwidth}
@{}}
\toprule
\textbf{Metric (Full name)} & \textbf{Formula / definition} \\
\midrule

\rowcolor{gray!15}
\multicolumn{2}{@{}l}{\emph{\textbf{Task completion and navigation}}} \\

\textbf{SR} (Success Rate)
&
\(\operatorname{SR}=N^{-1}\sum_{i=1}^{N}s_i\), where
\(s_i\in\{0,1\}\) indicates task success.
\\

\textbf{SPL} (Success weighted by Path Length)~\cite{spl_eval}
&
\(\operatorname{SPL}=N^{-1}\sum_{i=1}^{N}
s_i\ell_i/\max(p_i,\ell_i)\), where \(\ell_i\) and \(p_i\) are the
shortest and executed path lengths.
\\

\textbf{NE} (Navigation Error)~\cite{r2r}
&
\(\operatorname{NE}=N^{-1}\sum_{i=1}^{N}
d_{\mathcal G}(v_i^{\mathrm{stop}},v_i^{\mathrm{goal}})\), where
\(d_{\mathcal G}\) is the geodesic shortest-path distance.
\\

\textbf{nDTW} (normalized Dynamic Time Warping)~\cite{ndtw}
&
\(\operatorname{nDTW}_i=
\exp[-\operatorname{DTW}_{d_{\mathcal G}}
(\mathbf P_i^\star,\hat{\mathbf P}_i)/
(|\mathbf P_i^\star|d_{\mathrm{th}})]\);
report \(N^{-1}\sum_i\operatorname{nDTW}_i\).
\\

\midrule
\rowcolor{gray!15}
\multicolumn{2}{@{}l}{\emph{\textbf{Trajectory and localization}}} \\

\textbf{ADE} (Average Displacement Error)~\cite{argoverse,waymo}
&
\(\operatorname{ADE}=N^{-1}\sum_{i=1}^{N}T_i^{-1}
\sum_{t=1}^{T_i}
\|\hat{\mathbf x}_{i,t}-\mathbf x_{i,t}\|_2\).
\\

\textbf{FDE} (Final Displacement Error)~\cite{argoverse,waymo}
&
\(\operatorname{FDE}=N^{-1}\sum_{i=1}^{N}
\|\hat{\mathbf x}_{i,T_i}-\mathbf x_{i,T_i}\|_2\).
\\

\textbf{ATE} (Absolute Trajectory Error)~\cite{tum_rgbd}
&
After alignment by \(\mathbf S\in\mathrm{SE}(3)\),
\(\operatorname{ATE}=
[N^{-1}\sum_{i=1}^{N}
\|\operatorname{trans}(\mathbf T_i^{-1}
\mathbf S\hat{\mathbf T}_i)\|_2^2]^{1/2}\).
\\

\textbf{RPE} (Relative Pose Error)~\cite{tum_rgbd}
&
\(\operatorname{RPE}_{\mathrm{trans}}=
[(N-\Delta)^{-1}\sum_{i=1}^{N-\Delta}
\|\operatorname{trans}(\mathbf E_i)\|_2^2]^{1/2}\), where
\(\mathbf E_i=(\mathbf T_i^{-1}\mathbf T_{i+\Delta})^{-1}
(\hat{\mathbf T}_i^{-1}\hat{\mathbf T}_{i+\Delta})\).
\\

\midrule
\rowcolor{gray!15}
\multicolumn{2}{@{}l}{\emph{\textbf{Generation quality}}} \\

\textbf{FID} (Fr\'echet Inception Distance)~\cite{fid}
&
\(\operatorname{FID}=
\|\boldsymbol\mu_r-\boldsymbol\mu_g\|_2^2+
\operatorname{tr}[\boldsymbol\Sigma_r+\boldsymbol\Sigma_g-
2(\boldsymbol\Sigma_r^{1/2}\boldsymbol\Sigma_g
\boldsymbol\Sigma_r^{1/2})^{1/2}]\).
\\

\textbf{FVD} (Fr\'echet Video Distance)~\cite{fvd}
&
The same Fr\'echet form as FID, computed on video-feature distributions.
\\

\textbf{PSNR} (Peak Signal-to-Noise Ratio)
&
\(\operatorname{PSNR}=10\log_{10}
[v_{\max}^{2}/\operatorname{MSE}(\mathbf I,\hat{\mathbf I})]\),
where \(v_{\max}\) is the declared maximum pixel value.
\\

\textbf{SSIM} (Structural Similarity Index)~\cite{ssim}
&
\(\operatorname{SSIM}(\mathbf I,\hat{\mathbf I})=
[(2\mu_I\mu_{\hat I}+c_1)(2\sigma_{I\hat I}+c_2)]/
[(\mu_I^2+\mu_{\hat I}^2+c_1)
(\sigma_I^2+\sigma_{\hat I}^2+c_2)]\).
\\

\textbf{LPIPS} (Learned Perceptual Image Patch Similarity)~\cite{lpips}
&
\(\sum_l(H_lW_l)^{-1}\sum_{h,w}
\|\mathbf w_l\odot[
\hat{\boldsymbol\phi}_l(\mathbf I)_{h,w}-
\hat{\boldsymbol\phi}_l(\hat{\mathbf I})_{h,w}]
\|_2^2\).
\\

\midrule
\rowcolor{gray!15}
\multicolumn{2}{@{}l}{\emph{\textbf{Language and captioning}}} \\

\textbf{BLEU} (Bilingual Eval. Understudy)~\cite{bleu}
&
\(\operatorname{BLEU}=\operatorname{BP}
\exp(\sum_{n=1}^{M}w_n\log p_n)\), using modified \(n\)-gram precision
\(p_n\) and brevity penalty \(\operatorname{BP}\).
\\

\textbf{ROUGE-L} (Recall-Oriented Understudy for Gisting Eval.)%
~\cite{rouge}
&
\(\operatorname{ROUGE\text{-}L}=
(1+\beta^2)P_{\mathrm{LCS}}R_{\mathrm{LCS}}/
(R_{\mathrm{LCS}}+\beta^2P_{\mathrm{LCS}})\).
\\

\textbf{CIDEr} (Consensus-based Image Description Eval.)~\cite{cider}
&
\(\operatorname{CIDEr}=
\sum_{n=1}^{4}w_n|\mathcal R|^{-1}
\sum_{r\in\mathcal R}
\cos[\mathbf g_n(c),\mathbf g_n(r)]\), where
\(\mathbf g_n\) is a TF--IDF \(n\)-gram vector.
\\

\textbf{METEOR} (Metric for Eval. of Translation with Explicit
ORdering)~\cite{meteor}
&
\(\operatorname{METEOR}=F_{\mathrm{mean}}
(1-\operatorname{Penalty})\), where
\(F_{\mathrm{mean}}=PR/[\alpha P+(1-\alpha)R]\) and the penalty measures
alignment fragmentation.
\\

\midrule
\rowcolor{gray!15}
\multicolumn{2}{@{}l}{\emph{\textbf{Driving and RL aggregate}}} \\

\textbf{DS} (Driving Score)~\cite{carla,bench2drive}
&
\(\operatorname{DS}=N^{-1}\sum_{i=1}^{N}
\operatorname{RC}_i\operatorname{IS}_i\):
route-level products are averaged, rather than multiplying separate means.
\\

\textbf{PDMS} (Predictive Driver Model Score)~\cite{navsim}
&
\(q_i=\operatorname{NC}_i\operatorname{DAC}_i
(5\operatorname{TTC}_i+5\operatorname{EP}_i+2\operatorname{C}_i)/12\),
\(\operatorname{PDMS}=N^{-1}\sum_iq_i\);
TTC is a normalized subscore.
\\

\textbf{HNS} (Human Normalized Score)~\cite{atari}
&
\(\operatorname{HNS}_g=
(s_{\mathrm{agent},g}-s_{\mathrm{random},g})/
(s_{\mathrm{human},g}-s_{\mathrm{random},g})\),
computed per game before cross-game aggregation.
\\

\bottomrule
\end{tabular*}

\begin{tablenotes}[flushleft]
\scriptsize
\item \textit{Note:} $N$ is the number of evaluated episodes or samples; $T_i$ is the prediction horizon of sample $i$; $\hat{\mathbf{x}}_{i,t}$ and $\mathbf{x}_{i,t}$ are predicted and reference states; $\hat{\mathbf{P}}_i$ and $\mathbf{P}^{\star}_i$ are predicted/executed and reference paths; and $\mathcal{R}$ is a reference sentence set.
\end{tablenotes}

\end{threeparttable}
\end{table*}

%% file: final_chapters/tables/tab_unified_benchmark_v3.tex
\providecommand{\WAMCapChip}[3][15]{%
  \begingroup
  \setlength{\fboxsep}{0.5pt}%
  \colorbox{#2!#1}{\textcolor{#2}{\textbf{#3}}}%
  \endgroup
}

\begin{table*}[!t]
\centering
\scriptsize
\caption{
Summary of surveyed datasets, simulators, and benchmarks.
Columns report \textbf{Name}, \textbf{Reported Scale}, \textbf{Type}, and \textbf{Eval Metrics}.
Scales retain the units used by their primary sources and are not
directly comparable. Here, ``k'' and ``M'' denote thousand and million,
respectively. \textbf{N/A} denotes an open-ended simulator, whereas
``---'' indicates unavailable scale information.
}
\label{tab:unified-benchmarks}
\setlength{\tabcolsep}{2pt}
\setlength{\fboxsep}{1pt}
\setlength{\extrarowheight}{0pt}
\renewcommand{\Sim}{\colorbox{simblue!18}{\textcolor{simblue}{\textbf{Sim}}}}
\renewcommand{\Real}{\colorbox{realgreen!18}{\textcolor{realgreen}{\textbf{Real}}}}
\renewcommand{\Mixed}{\colorbox{mixedpurple!18}{\textcolor{mixedpurple}{\textbf{Mixed}}}}
\renewcommand{\arraystretch}{0.5}
\begin{tabular*}{\textwidth}{@{\extracolsep{\fill}}
  l c c l
  l c c l
  @{}}
\toprule
\textbf{Name} & \textbf{Reported Scale} & \textbf{Type} & \textbf{Eval Metrics} &
\textbf{Name} & \textbf{Reported Scale} & \textbf{Type} & \textbf{Eval Metrics} \\
\midrule

\rowcolor{gray!15}
\multicolumn{8}{@{}l}{
  \emph{\textbf{Robot demonstration and human video corpora}}
} \\

DROID~\cite{droid}
& 76k~trajectories
& \Real
& SR, FVD, FID
&
Ego4D~\cite{ego4d}
& 3.67k~hours
& \Real
& SR, ATE, RPE
\\

BridgeData V2~\cite{bridgev2}
& 60{,}096~trajectories
& \Real
& SR, Accuracy
&
EgoDex~\cite{egodex}
& \mbox{829~hours / 194~tasks}
& \Real
& SR (downstream)
\\

Open X-Embodiment~\cite{oxe}
& \mbox{1M+~trajectories}
& \Mixed
& SR, FVD, LPIPS
&
Kinetics-700~\cite{kinetics}
& 700~classes
& \Real
& SR (downstream)
\\

AgiBot-World~\cite{agibotworld}
& 1M+~trajectories
& \Real
& SR, FVD, LPIPS
&
HowTo100M~\cite{howto100m}
& 136M~clips
& \Real
& SR (downstream)
\\

SSv2 V2~\cite{ssv2}
& 220k~clips
& \Real
& SR (downstream)
&
&
&
&
\\

\midrule

\rowcolor{gray!15}
\multicolumn{8}{@{}l}{
  \emph{\textbf{Manipulation task suites and benchmarks}}
} \\

LIBERO~\cite{libero}
& 130~tasks
& \Sim
& SR, FVD, LPIPS, PSNR
&
MimicGen~\cite{mimicgen}
& \mbox{18~tasks / 50k+~demos}
& \Sim
& SR, FVD
\\

CALVIN~\cite{calvin}
& 34~tasks
& \Sim
& Avg.\ Len., SR, FID
&
RLBench~\cite{rlbench}
& 100+~tasks
& \Sim
& SR
\\

MetaWorld~\cite{metaworld}
& 50~tasks
& \Sim
& SR, Norm.\ Score
&
ManiSkill (1/2/3)~\cite{maniskill}
& 20+~tasks
& \Sim
& SR, Norm.\ Score
\\

RoboTwin / 2.0~\cite{robotwin,robotwin2}
& 50+~tasks
& \Mixed
& SR, PSNR, SSIM
&
SimplerEnv~\cite{simplerenv}
& ---
& \Sim
& SR, Accuracy
\\

RoboCasa~\cite{robocasa}
& \mbox{100~tasks / 120~scenes}
& \Sim
& SR, FVD, SSIM
&
Push-T~\cite{diffusion_policy}
& 1~task
& \Sim
& IoU, SR
\\

RoboDojo~\cite{robodojo}
& 60~tasks
& \Mixed
& SR, Avg.\ Score
&
&
&
&
\\

\midrule

\rowcolor{gray!15}
\multicolumn{8}{@{}l}{
  \emph{\textbf{Autonomous driving datasets and simulators}}
} \\

nuScenes~\cite{nuscenes}
& 1k~scenes
& \Real
& Coll., L2, mIoU, FVD
&
Waymo Open Motion~\cite{waymo}
& 103k~segments
& \Real
& ADE, FDE, Miss Rate
\\

nuPlan~\cite{nuplan}
& 1.3k~hours
& \Real
& PDMS, Comfort, TTC
&
OpenDV-2K~\cite{opendv}
& 2{,}059~hours
& \Real
& FID, FVD, LPIPS
\\

NAVSIM v1/v2~\cite{navsim,navsimv2}
& ---
& \Mixed
& PDMS, EPDMS, NC
&
BDD100K~\cite{bdd100k}
& 100k~videos
& \Real
& Accuracy, mAP
\\

Bench2Drive~\cite{bench2drive}
& \mbox{220~routes / 12~towns}
& \Sim
& DS, RC, IS, L2
&
KITTI~\cite{kitti}
& 6~hours
& \Real
& FID, FVD
\\

CARLA~\cite{carla}
& N/A
& \Sim
& DS, IS, RC, Coll.
&
Argoverse~\cite{argoverse}
& 327.8k~sequences
& \Real
& ADE, FDE
\\

OpenScene~\cite{openscene}
& 120+~hours
& \Real
& mIoU, Coll.\ Rate
&
ZOD~\cite{zod}
& \mbox{1.5k~scenes}
& \Real
& mAP
\\

\midrule

\rowcolor{gray!15}
\multicolumn{8}{@{}l}{
  \emph{\textbf{Navigation environments, datasets, and benchmarks}}
} \\

Matterport3D~\cite{matterport3d}
& 90~buildings
& \Real
& SR, SPL, NE, nDTW
&
RECON~\cite{recon}
& ---
& \Real
& SR
\\

R2R / R2R-CE~\cite{r2r,r2r_ce}
& 7.2k~paths
& \Sim
& SR, SPL, NE, Oracle
&
TartanDrive~\cite{tartandrive}
& ---
& \Real
& SR
\\

RxR / RxR-CE~\cite{rxr}
& 16.5k~paths
& \Sim
& SR, SPL, NE
&
SCAND~\cite{scand}
& ---
& \Real
& SR
\\

Habitat~\cite{habitat}
& N/A
& \Sim
& SR, SPL, ATE, RPE
&
ScaleVLN~\cite{scalevln}
& 4.9M~trajectories
& \Sim
& SR, SPL
\\

\midrule

\rowcolor{gray!15}
\multicolumn{8}{@{}l}{
  \emph{\textbf{Foundational reinforcement-learning benchmarks}}
} \\

DMControl~\cite{dmcontrol}
& 28~tasks
& \Sim
& Return, Norm.\ Score
&
Craftax~\cite{craftx}
& 1~game
& \Sim
& Return, Reward
\\

Atari (ALE)~\cite{atari}
& 55~games
& \Sim
& HNS, IQM, Mean/Md.
&
DMLab~\cite{dmlab}
& ---
& \Sim
& Return
\\

Atari 100k~\cite{atari100k}
& \mbox{26~games / 100k~steps}
& \Sim
& HNS, IQM
&
BSuite~\cite{bsuite}
& 23~tasks
& \Sim
& Return
\\

Procgen~\cite{procgen}
& 16~games
& \Sim
& Return, Reward
&
FMB~\cite{fmb}
& ---
& \Real
& SR, Cycle Time
\\
\bottomrule

\end{tabular*}
\end{table*}

%% file: final_chapters/Sec-VI.tex
\section{Open Challenges and Future Directions}
\label{sec:challenges}

WAMs integrate perception, prediction, and action generation into a unified framework, 
posing a challenging multi-objective optimization problem for their design and training.
Their development currently faces four major bottlenecks: action grounding, spatial consistency, closed-loop policy improvement, and real-time inference.

\subsection{Decoupling Motion Intent and Action Alignment}
\label{subsec:wam_challenges_1}

Compared with conventional VLA policies~\cite{pi0,openvla}, WAMs~\cite{cosmos_policy,dreamzero,uwm} decouple task-level motion reasoning from low-level robot control. This separation raises two central challenges under the imbalance between visual and action data: staged optimization of motion intent (`what to happen') learning and action alignment (`how to act'), and preservation of pretrained visual priors.

\subsubsection{Stage-wise Motion-to-Action Alignment}
IDM-style WAMs~\cite{dreamzero,lingbotva,gigaworld_policy} are often well suited to staged optimization. A common strategy is to pretrain video foundation models~\cite{cosmos,wan} on large-scale visual data to capture spatiotemporal dynamics, and then adapt them with embodied data to align predicted motion to executable actions.

\subsubsection{Prior Preservation during Action Alignment}
A second challenge is preserving pretrained visual priors during action alignment. Parameter freezing, action-alignment layers, and residual action heads can confine control-specific updates. Additionally,  stronger anti-forgetting mechanisms may help retain the broad knowledge acquired during large-scale pre-training.

\subsection{World--Action Factorization}
\label{subsec:wam_challenges_factorization}

From the agent–environment perspective widely studied in reinforcement learning~\cite{sutton}, world prediction and action selection are related but fundamentally different problems. The former models action-conditioned state transitions and prioritizes predictive fidelity, whereas the latter maps states to actions and maximizes cumulative rewards. Their joint trajectory distribution naturally factorizes as
\begin{equation}
\begin{aligned}
  p^{\pi}_{\mu}(a_t, s_{t+1}, \dots, a_{T-1}, s_T \mid s_t)
  &= \\
  \prod_{k=t}^{T-1}
  {\underbrace {\textstyle p_\pi(a_{k} \mid s_k)}_{\text{ agent }} } \ & {\underbrace {\textstyle p_\mu(s_{k+1} \mid s_k, a_{k})}_{\text{ universe }} }.
  \label{eq:trajectory}
\end{aligned}
\end{equation}

\subsubsection{Decoupled Learning Objectives}
World--action factorization separately trains the world and agent models according to their natural objectives~\cite{xing2026CAM,xing2025critiques}. The world model focuses on accurate simulation, while the agent model optimizes action selection. By contrast, joint optimization may encourage a monolithic WAM to generate overly optimistic states that are easily exploited by its policy, potentially degrading real-world performance.

\subsubsection{Decoupled Learning and Inference}
Separating the two models also enables continual training and flexible inference. The world model can serve as a simulator for offline reinforcement learning, runtime planning, and model predictive control. Although this direction has shown promising results~\cite{xing2026CAM}, controlling model bias and coordinating the two components remain important challenges.


\subsection{Spatial and Multi-View Consistency}
\label{subsec:wam_challenges_2}
Robotic control requires more than single-view 2D prediction. 
Current video backbones often struggle to enforce geometric constraints, maintain object permanence, and ensure cross-view consistency. These limitations can degrade state estimation, planning, and inverse-dynamics decoding in manipulation and autonomous driving.

\subsubsection{Explicit 3D Spatial Representations}
Structured 3D prediction offers a more control-relevant alternative to independent view generation. Representative choices include point clouds~\cite{pointworld}, occupancy fields~\cite{occupancy}, and 3D Gaussian representations~\cite{3dgs}. The main challenge is to integrate them with scalable generative backbones and efficient action decoding.

\subsubsection{Egocentric Observations}
Egocentric videos provide rich cues about manipulation intent and hand-object dynamics~\cite{egodex,egoscale}. With fixed inter-camera extrinsics, it also tends to offer stronger multi-view consistency~\cite{SCVC}.



\subsection{Long-Horizon Memory}
\label{subsec:wam_challenges_memory}

Long-horizon tasks require agents to retain task progress, past interactions, and occluded information. Simply extending the observation window is computationally expensive and introduces redundant visual context, motivating compact and selective memory mechanisms.

\subsubsection{Model-Integrated Memory Architectures}
MEM~\cite{torne2026mem} combines video-based short-term memory with text-based long-term memory. EventVLA~\cite{yang2026eventvla} selectively stores task-critical visual events, while MemoryVAM~\cite{jiang2026memoryvam} injects episodic memory into both video prediction and action decoding. However, such methods often require architecture-specific training.

\subsubsection{Memory-Augmented Agentic Harnesses}
Harness VLA~\cite{zhang2026harnessvla} externalizes memory management by combining a frozen VLA with an agentic planner that maintains execution traces, success rules, and failure models. This modular design supports planning, verification, and failure recovery without modifying the underlying policy, offering a promising direction for scalable long-horizon memory.


\subsection{Neural Simulators and Closed-Loop Policy Learning}
\label{subsec:wam_challenges_3}
Imitation learning~\cite{diffusion_policy} can approximate expert state-action distributions, but its closed-loop correction capability remains limited and it relies heavily on high-quality expert demonstrations. Low-quality or failed demonstrations are typically discarded, and training skilled teleoperators is costly. Moreover, current imitation learning methods have limited ability to improve from failure, motivating closed-loop policy learning~\cite{pi_rl}.

\subsubsection{Neural Simulators}
Neural simulators based on RSSM-style models~\cite{iws,td_mpc} learn action-conditioned dynamics that serve as predictive environments for policy evaluation and improvement~\cite{Motus2}, although prediction errors accumulated during imagination can introduce model bias and complicate long-horizon policy optimization~\cite{trust}.

\subsubsection{Closed-Loop Policy Learning}
Reinforcement learning in real systems is often constrained by sample inefficiency, while simulation-based learning suffers from the sim2real gap. Neural simulators enable policy optimization through imagined interactions using task rewards~\cite{dreamer,td_mpc}, goal-based objectives~\cite{lewm,dino_wm}, or learned reward models~\cite{pi0_6,gigabrain0_5}, where long-horizon credit assignment is critical for improving policies from predicted futures~\cite{wovr}.
Self-supervised reinforcement learning (SSL-RL)~\cite{Benjamin_2019,eysenbach2022contrastive} may replace the IDM-based action expert in dual systems (Section~\ref{sec:wam_architecture}), but reward-free and offline learning remain key challenges at scale.


\subsection{Inference Latency and Computational Efficiency}
\label{subsec:wam_challenges_4}
As WAMs scale, inference latency becomes a control bottleneck. Multi-step sampling, rollout generation, and action decoding can delay feedback in dynamic or contact-rich systems. Efficiency is therefore a control requirement, not only an implementation detail.

\subsubsection{Few-Step Action Generation}
Few-step generation is essential for real-time WAMs~\cite{SANTS}. Consistency models~\cite{cm}, mean-flow~\cite{mean_flow}, rectified-flow~\cite{rectified_flow}, and related methods can reduce sampling cost, but these acceleration methods must preserve action-conditioned transitions, temporal consistency, and control-relevant state information.

\subsubsection{Deployment Optimization}
System-level optimizations such as quantization, pruning, operator fusion, batching, memory-layout optimization~\cite{PhyAI}, distillation~\cite{dmd,dmd2}, and KV-cache reuse~\cite{dreamzero} are essential. Overall efficiency also depends on rollout horizon, imagined-trajectory count, replanning frequency, and action-decoding cost, which should be reduced without compromising control fidelity~\cite{FBFM}.

%% file: final_chapters/Sec-VII.tex
\section{Conclusion}
\label{sec:conclusion}
World-Action Models (WAMs) aim to close the gap between predictive world modeling and executable robot control. By conditioning future-state prediction on actions and using the resulting predictions for action selection, policy learning, or inverse-dynamics decoding, WAMs extend beyond VLA policies. This survey reviewed this emerging class from a robotics perspective, organized its main design choices into a unified taxonomy, and related representative methods to manipulation, navigation, autonomous driving, and generalist embodied robotics.
Useful predictions must be grounded in robot actions, preserve spatial and temporal consistency, expose uncertainty, support closed-loop improvement, and satisfy real-time constraints. Progress will therefore depend on tighter coupling among multimodal and 3D world representations, action-alignment mechanisms, neural simulation, reinforcement learning, and real-robot feedback, together with benchmarks that measure task success, safety, robustness, and embodiment transfer.